\documentclass[lettersize,10pt,journal]{IEEEtran}

\usepackage{color,array}
\ifCLASSINFOpdf
\else
\fi
\usepackage{array}
\usepackage[hang]{footmisc}

\usepackage[subrefformat=parens,labelformat=parens,caption=false,font=footnotesize]{subfig}
\usepackage{amsmath}
\usepackage[nolist,nohyperlinks]{acronym}
\usepackage{amsmath}
\DeclareMathOperator*{\argmax}{arg\,max}

	\begin{acronym}
    \acro{4G}{fourth generation}
    \acro{5G}{fifth generation}
    \acro{AoA}{angle of arrival}
   	\acro{AoD}{angle of departure}
   	\acro{AP}{access point}
    \acro{BAI}{best arm identification}
    \acro{BCRLB}{Bayesian CRLB}
    \acro{BS}{base stations}
	\acro{CDF}{cumulative density function}
    \acro{CF}{closed-form}
    \acro{CRLB}{Cramer-Rao lower bound}
    \acro{ECDF}{empirical cumulative distribution function}
    \acro{EI}{Exponential Integral}
    \acro{eMBB}{enhanced mobile broadband}
    \acro{FIM}{Fisher Information Matrix}
    \acro{GoF}{goodness-of-fit}
    \acro{GPS}{global positioning system}
    \acro{GNSS}{global navigation satellite system}
    \acro{HetNets}{heterogeneous networks}
    \acro{LOS}{line of sight}
    \acro{MAB}{multi-armed bandit}
    \acro{MBS}{macro base station}
    \acro{MEC}{mobile-edge computing}
    \acro{MIMO}{multiple input multiple output}
    \acro{mm-wave}{millimeter wave}
    \acro{mMTC}{massive machine-type communications}
    \acro{MS}{mobile station}
    \acro{MVUE}{minimum-variance unbiased estimator}
    \acro{NLOS}{non line-of-sight}
    \acro{OFDM}{orthogonal frequency division multiplexing}
    \acro{PDF}{probability density function}
    \acro{PGF}{probability generating functional}
    \acro{PLCP}{Poisson line Cox process}
    \acro{PLT}{Poisson line tessellation}
    \acro{PLP}{Poisson line process}
    \acro{PPP}{Poisson point process}
    \acro{PV}{Poisson-Voronoi}
    \acro{QoS}{quality of service}
    \acro{RAT}{radio access technique}
    \acro{RL}{reinforcement-learning}
    \acro{RE}{rapid exploration}
    \acro{RSSI}{received signal-strength indicator}
    \acro{LRT}{likelihood ratio test}
    \acro{BS}{base station}
   	\acro{SINR}{signal to interference plus noise ratio}
    \acro{SNR}{signal to noise ratio}
    \acro{SH}{sequential halving}
    \acro{SR}{successive rejects}
    \acro{TS}{Thompson Sampling}
    \acro{TS-CD}{TS with change-detection}
    \acro{KS}{Kolmogorov-Smirnov}
    \acro{UCB}{upper confidence bound}
	\acro{ULA}{uniform linear array}
	\acro{UE}{user equipment}
 	\acro{URLLC}{ultra-reliable low-latency communications}
    \acro{V2V}{vehicle-to-vehicle}    
\end{acronym}
\usepackage{caption}
\usepackage{graphicx,wrapfig,lipsum}
\usepackage{rotating}
\usepackage{amsmath, amssymb, amsthm}
\usepackage{algorithmic}
\usepackage{algorithm}
\usepackage{stmaryrd}
\usepackage{tikz}
\usepackage{verbatim}
\usepackage{url}
\usepackage[utf8]{inputenc}
\usepackage[english]{babel}
\newtheorem{theorem}{Theorem}[]
\newtheorem{corollary}{Corollary}[]

\newtheorem{lemma}[]{Lemma}
\usepackage{multicol}
\usepackage{xr-hyper}
\newtheorem{remark}{Remark}[]
\newtheorem{assumption}{Assumption}[]
\usepackage{amsmath,amssymb,xcolor,mathtools}
\usepackage{scalerel}
\usepackage{multicol}
\usepackage{multirow}
\usepackage{setspace}

\usepackage[noadjust]{cite}
\usepackage{booktabs}
\newcommand{\cA}{\mathcal{A}}

\newcommand{\cN}{\mathcal{N}}

\newcommand{\cP}{\mathcal{P}}

\newcommand{\bE}{\mathbb{E}}
\newcommand{\bP}{\mathbb{P}}
\newcommand{\N}{\mathbb{N}}

\usepackage{bbm}
\usepackage[normalem]{ulem}
\usepackage{color}
\usepackage{dsfont}
\usepackage{bm}
\usepackage{setspace}
\usetikzlibrary{automata}
\usetikzlibrary{arrows}
\usetikzlibrary{shapes.geometric, arrows}
\usepackage[]{footmisc}
\usetikzlibrary{positioning}
\usetikzlibrary{calc}
\usetikzlibrary{arrows}
\allowdisplaybreaks

\newcommand\blfootnote[1]{%
  \begingroup
  \renewcommand\thefootnote{}\footnote{#1}%
  \addtocounter{footnote}{-1}%
  \endgroup
}
\usepackage{mathtools}

\usepackage{xcolor}
\usepackage{scalerel}
\usepackage{float}
\usepackage{tikz}
\usetikzlibrary{svg.path}
\usepackage{tikz,xcolor,hyperref}

\definecolor{lime}{HTML}{A6CE39}
\DeclareRobustCommand{\orcidicon}{%
	\begin{tikzpicture}
	\draw[lime, fill=lime] (0,0) 
	circle [radius=0.16] 
	node[white] {{\fontfamily{qag}\selectfont \tiny ID}};
	\draw[white, fill=white] (-0.0625,0.095) 
	circle [radius=0.007];
	\end{tikzpicture}
	\hspace{-2mm}
}

\foreach \x in {A, ..., Z}{%
	\expandafter\xdef\csname orcid\x\endcsname{\noexpand\href{https://orcid.org/\csname orcidauthor\x\endcsname}{\noexpand\orcidicon}}
}

\begin{document}

	\title{\fontsize{22.8}{27.6}\selectfont An Algorithm for Fixed Budget Best Arm Identification with Combinatorial Exploration
    }
	\author{Siddhartha Parupudi and Gourab Ghatak \thanks{The authors are with the Department of Electrical Engineering, IIT Delhi, Hauz Khas, India 110016. Email:P.Siddhartha.ee121@ee.iitd.ac.in, gghatak@ee.iitd.ac.in.}
  }
\setcounter{page}{1}
	\maketitle

\begin{abstract}
We consider the best arm identification (BAI) problem in the $K-$armed bandit framework with a modification - the agent is allowed to play a subset of arms at each time slot instead of one arm. Consequently, the agent observes the sample average of the rewards of the arms that constitute the probed subset. Several trade-offs arise here - e.g., sampling a larger number of arms together results in a wider view of the environment, while sampling fewer arms enhances the information about individual reward distributions. Furthermore, grouping a large number of suboptimal arms together albeit reduces the variance of the reward of the group, it may enhance the group mean to make it close to that containing the optimal arm. To solve this problem, we propose an algorithm that constructs $\log_2 K$ groups and performs {a likelihood ratio test (LRT) with the worst case prior} to detect the presence of the best arm in each of these groups. Then a Hamming decoding procedure determines the unique best arm. We derive an upper bound for the error probability of the proposed algorithm based on a new hardness parameter $H_4$. Finally, we demonstrate the efficacy of the proposed algorithm by employing it in two case studies - one on jammer waveform selection and the other on active radar channel detection. Importantly, we highlight the cases in which the proposed algorithm outperforms the competing algorithms in addition to the cases where its performance deteriorates.
\end{abstract}

%\begin{IEEEImpStatement}
%Our work will find applications in problems involving resource allocation. For example, in wireless communications, a transmitter may distribute its transmit power among multiple available channels or multiple antennas and only get the total received power as feedback. Eventually, the transmitter must identify the best channel to continue data transmission. Similarly, a trader may want to distribute an exploratory capital among different investment instruments while receiving the total profit/loss as a feedback. Here after a fixed budget, the trader has to identify the most profitable investment instrument for an eventual bulk investment. In spite of obvious applications, to the best of our knowledge, this setting has not been studied in literature previously. Our work presents the first steps in this regard and it is envisaged that our study will open new research directions. In particular, a key technical challenge is to derive the lower bound of the error probability which remains an open information theoretic problem.
%\end{IEEEImpStatement}

\section{Introduction}
A \ac{MAB} framework $\nu_{[K]}$ ($[K] \doteq \{1, 2, \ldots, K\}$), consists of a collection of $K$ arms, where each arm $\nu_a \; (1\leq a\leq K)$ is a probability distribution, typically on but not limited to $\mathbb{R}$ with an expectation $\mu_a$. We denote the sample space of $\nu$ by $\mathcal{N}$. In the single-play setting, at each discrete time slot $t \in \mathbb{Z}^+$, an agent chooses an arm $A_t\in\{1,\dots,K\}$ and receives an independent sample $X_t$ from the corresponding arm $\nu_{A_t}$. We denote by $\bP_\nu$ (resp. $\bE_\nu$) the probability law (resp. expectation) of the process $\{X_t\}$. The \ac{BAI} is a specific problem setting in the \ac{MAB} model, where the agent is required to identify $a^* = \argmax_a \mu_a$. The performance of a policy which recommends the arm $\hat{a}$ in the \ac{BAI} setting is determined by either the probability of error in determining the best arm or in terms of the number of samples it needs for recommendation. This is in contrary to the classical {\it regret} framework, wherein, the agent has to optimize the exploration-exploitation dilemma during the action selection. {To elaborate, recall that any algorithm that is regret optimal may not guarantee an exponentially bounded error probability in the \ac{BAI} setting~\cite{bubeck2009pure}. Similarly, any algorithm that achieves exponential error probability for the \ac{BAI} setting may not achieve sub-linear regret in the classical regret setting. Consequently, combinatorial bandit algorithms tailored to the regret setting (e.g., see \cite{kolla2018collaborative, gentile2014online}) cannot be effectively applied for the \ac{BAI} problem.}

Two different variants of the \ac{BAI} problem are studied in literature. In the \emph{fixed-confidence} variant, for a given error probability $\delta$, the agent attempts to minimize the expectation of the number of plays, called sample complexity, $\tau_\delta$, before recommending the best arm. A strategy $\cA(\delta)$ is called \emph{$\delta$-PAC} if, for every choice of $\nu$ in $\mathcal{N}$, $\bP_\nu(\hat{a} = a^*)\geq 1-\delta$. On the contrary, in the \emph{fixed-budget} variant, the number of plays $T \in \N$ is fixed, while the goal is to choose the sampling and recommendation rules so as to minimize the error probability of recommending the best arm $p_T(\nu) := \bP_\nu(\hat{a} \neq a^*)$. In the fixed-budget setting, a family of strategies $\cA(T)$ is called \emph{consistent} if, for every choice of $\nu \in \mathcal{N}$, $p_T(\nu)$ tends to zero when $T$ tends to infinity. In what follows, we adopt the notation $p_{\rm D}^A(T)$ to denote the probability of error of algorithm $A$ in a budget $T$, where the reward distribution is $D$.

\subsection{Related Works}
    For the fixed budget setting, Audibert et al.~\cite{audibert2010best} proposed the \texttt{UCB-E} and \texttt{Successive Rejects} (SR) algorithms and proved their optimality up to logarithmic factors. Specifically, the upper bound on the probability of error for their proposed algorithms is
%\begin{align}
    $p(T) \leq \frac{K (K-1)}{2} \exp\left(- \frac{T - K}{\log_2 (K) H_2}\right),$
%\end{align}
where $H_2$ is called the hardness of the problem that depends on the specific instance of $\nu \in \mathcal{N}$ (to be discussed in detail soon). In a remarkable result, Carpentier \& Locatelli~\cite{carpentier2016tight} proved the following lower bound for the probability of error in the fixed-budget setting:
%\begin{align}
       $p(T) \gtrsim \exp\left(-\frac{T}{\log(K) H}\right),$
%\end{align}
where $H$ is another variant of the hardness parameter. This disproved a long-standing assumption that there must exist an algorithm for this problem whose probability of error is upper bound by $\exp(-T/H)$. This established a key difference between the fixed-confidence and the fixed-budget settings. Following this line of work, Karnin~\cite{karnin2013almost} proposed the \texttt{SEQUENTIAL HALVING} ({SH}) algorithm and proved it to be almost optimal for BAI problems. This is achieved by eliminating half of the surviving arms with the worst estimates in each round. \ac{BAI} problem has also been analyzed using other variants of the \ac{UCB} algorithm (e.g., LUCB of \cite{LUCB}) which are not based on eliminations. Most of the experimental results of these algorithms are present for bounded distributions, that are in fact particular examples of distributions with sub-Gaussian tails. Since then, the fixed budget pure exploration problem has been studied in a wide variety of contexts, e.g., for linear bandits~\cite{jedra2020optimal}, minimax optimality~\cite{yang2022minimax}, spectral bandits~\cite{kocak2020best}, risk-averse bandits~\cite{kagrecha2022statistically}, and bandits with mis-specified linear models~\cite{alieva2021robust}. {From a fixed confidence point of view, Jourdan {\it et al.}~\cite{jourdan2021efficient} have studied a combinatorial pure-exploration problem with semi-bandit feedback. This is a simpler version than our setting where we assume that the player can only access a function of the rewards (specifically, sample mean) from the sampled arms, thereby abstracting arm-specific observations.} More recently, large deviation perspective was studied in~\cite{wang2024best}, cost-aware BAI in \cite{qin2025cost} and arm erasures in \cite{reddy2024best}. Thompson sampling has also been employed to form the {\it best challenger} rule to improve the efficiency of \ac{BAI}~\cite{lee2024thompson}. Interesting new directions include combining the regret setting with the BAI setting, e.g., see~\cite{zhang2024fast, qin2024optimizing}. { On the same lines, the top-$K$ subset selection with linear space and non-linear feedback has been investigated~\cite{agarwal2021stochastic, agarwal2022stochastic}}. Finally, assuming a distribution of $\nu$ over $\mathcal{N}$, researchers have studied rate-optimal Bayesian regret in \ac{BAI}~\cite{komiyama2024rate}.

\subsection{Motivation and Contributions}
In all of the above works, the agent is restricted to sample a single arm from $\nu_{[K]}$ at each time slot. We propose that in case this assumption is relaxed, i.e., if the agent is allowed to select a subset of arms, it may enhance the efficiency of \ac{BAI}, i.e., the error probability may reduce. Naturally, the case where the selection of a subset of arms leads to the agent having access to individual reward samples, is trivial - the optimal action here is to select all the arms. Hence, we focus on a setting where on the selection of a subset of the arms, the agent receives a single reward that is a function of the individual rewards of the selected arms. In this work, we assume this function to be the sample average of rewards of the selected arms. {It must be noted that other combined reward functions such as the maximum or minimum of the sample rewards can be more applicable for several use-cases. Based on our first attempt with the simple case of sample average of rewards, extension of the framework to more general class of combination functions is indeed an open problem and an interesting direction of research which we are currently investigating.} We make the following contributions.
\begin{itemize}
    \item {\bf Combinatorial Exploration in BAI:} {We introduce a novel fixed-budget \ac{BAI} framework where the agent selects subsets of arms instead of individual arms while receiving a sample average rewards of the arms that are sampled (e.g., unlike \cite{jourdan2021efficient}), enabling a trade-off between broader exploration and fine-grained reward estimation, a setting previously unexplored in the literature for non-trivial rewards (e.g., unlike \cite{ghatak2024best} where the rewards of sub-optimal arms were constant 0).}
    \item {\bf Algorithm and Theoretical Guarantees:}  We propose an innovative arm grouping strategy using Hamming codes, allowing efficient detection of the best arm through \ac{LRT} and a decoding procedure. Additionally, we introduce a new hardness parameter, $H_4$ and derive an upper bound on the error probability. Based on this we introduce the algorithm \ac{RE} for \ac{BAI} with combinatorial pulls.
    \item {\bf Empirical Analysis and Case Studies:} We conduct numerical experiments to identify the conditions under which our proposed algorithm outperforms state-of-the-art single-play algorithms. { In particular a detailed discussion on the hardness parameters is presented to outline the relative complexity of different algorithms. Finally, we employ the proposed algorithm in two real-world case studies - one on jammer waveform selection and the second a data-driven radar detection problem. We discuss cases where our method under performs, providing insights into its limitations.}
\end{itemize}

\section{Preliminaries}
Let $(\mu_{[1]}, \mu_{[2]}, \dots, \mu_{[K]})$ denote the ordered $K$-tuple of $(\mu_1, \mu_2, \dots, \mu_K)$ sorted in decreasing order. We define the sub-optimality gaps as
    {
\begin{align}\label{eqn:sub-optimality_gaps}
    \Delta_a &= \mu_{[1]} - \mu_a \ \ \text{for  } a \neq a^* = \argmax \mu_a, \\
    \Delta_{a^*} &= \min_{a \neq a^*} \Delta_a \ \ \ \text{for } a = a^* .
\end{align}
Note that $\Delta_{a^*}$ is not the sub-optimality gap of the optimal arm with itself (which is in fact 0), but rather the minimum sub-optimality gap in the system. Due to our ordering notation, $\Delta_{a^*} = \mu_{[1]} - \mu_{[2]}$. Next, we sort the sub-optimality gaps in increasing order, giving rise to 
\begin{equation}
    \Delta_{\min} = \Delta_{[1]} = \Delta_{[2]} \leq \Delta_{[3]} \leq \dots \leq \Delta_{[K]} = \Delta_{\max}.
\end{equation}
This is because we define the sub-optimality gap for the optimal arm as its distance to its closest competitor, or the smallest sub-optimality gap of the sub-optimal arms. Since in our framework, there is exactly one optimal/best arm, coincidentally the values of $\Delta_{a^*}$, $\Delta_{[1]}$, and $\Delta_{\min}$ are same. This will not be the case of a scenario with multiple optimal arms, however, such a scenario is beyond the scope of this work. Furthermore, note that the choice of $\Delta_{[1]} = \Delta_{\min}$ represents a worst-case scenario, since the lower the sub-optimality gap, the worse will the performance of an algorithm.} 
We now define a few relevant {\it hardness terms} below. Each of the hardness terms is a function of the specific $\nu \in \mathcal{N}$, hence we drop this parametrization unless needed.
\begin{align}
  H_1 &\triangleq \sum_{i \in \{1, 2, \dots
    K\}}\frac{1}{\Delta_i^2}, %\label{def:H_1}, \nonumber \\
    &H_2 \triangleq \max_{i \in \{2,\dots,K\}}\frac{i}{(\mu_{[1]}-\mu_{[i]})^2}, \nonumber \\
    H_3 &\triangleq  \frac{K}{\Delta_{[1]}^2}, \;\; \text{ and } 
    &H_4 \triangleq  \frac{1}{(\Delta_{[1]} + \Delta_{[K]})^2}.     \nonumber
\end{align}
The term $H_4$ is introduced by us in this paper for the first time, and we will need it to analyze our algorithm. The following inequalities hold true for the hardness terms
\begin{align}
  H_2 \leq H_1 \leq \log(2K) H_2, \nonumber \\
  H_2 \leq H_1 \leq H_3, \text{ and }\quad
  4H_4 \leq H_1 \leq  4KH_4.\nonumber
\end{align}
Let us first review the results for bounded rewards for the existing algorithms.
\subsection{Bounded Rewards}
\begin{lemma}
    For the uniform allocation strategy, the error probability is bounded as
\begin{align}\label{eqn:UE_bounded}
    p_{\rm U}^{\rm{UE}}(T) \leq (K-1)\exp\bigg(\frac{-T}{2H_3}\bigg). 
\end{align}
\end{lemma}
\begin{IEEEproof}
    Please see Appendix~\ref{app:UE}.
\end{IEEEproof}
\begin{lemma}
For rewards in $[0,1]$, the upper bound on the error probability of SR~\cite{audibert2010best} and SH~\cite{Sequential-halving} are
\begin{align}
    p_{\rm U}^{\mathsf{SR}} \leq \frac{K(K-1)}{2}\exp{\bigg(- \frac{T-K}{\log(K)H_2}\bigg)}, \label{eqn:SR_bounded}\\
    p_{\rm U}^{\mathsf{SH}}(T) \leq 3 \log_2(K) \exp{\bigg(-\frac{T}{8H_2\log_2(K)} \bigg)}. \label{eqn:SH_bounded}
\end{align}
\end{lemma}
\begin{remark}
The SR algorithm is better in terms of lower error probability for large budgets, but the SH algorithm scales better with $K$ for a fixed budget. Furthermore, when $\Delta_{[2]} = \Delta_{[K]}$, i.e., all the suboptimal arms have the same suboptimality gap, $H_1 = H_2 = H_3 = 4KH_4$. In this case, the uniform allocation strategy is moderately better than the other algorithms, however, when $\Delta_{[1]} = \Delta_{[2]}$ is small and all other suboptimality gaps are large, SR and SH are significantly better than the non-adaptive uniform exploration strategy.
\end{remark}

\subsection{Gaussian Rewards}
Although not explicitly reported in literature, in case the reward distribution is Gaussian, the probabilities of error are calculated using the bounds on the Gaussian $Q$-function rather than Hoeffding's inequality, and are presented below. We omit the proofs since they follows in the same manner as the bounded rewards case.
\begin{lemma}
The probability of error for fixed budget BAI for Gaussian rewards with fixed variance $\sigma^2$ for each of the three algorithms, respectively, are
    \begin{align}
    \label{eqn:UE_Gaussian}
    p_{\rm G}^{\mathsf{UE}}(T)\leq &(K-1)\sqrt{\frac{H_3\sigma^2}{\pi T}} \exp\bigg(\frac{-T}{4H_3\sigma^2}\bigg), \\
     p_{\rm G}^{\mathsf{SR}}(T) \leq &\frac{K(K-1)}{2} \sqrt{\frac{H_2\sigma^2\log(K)}{2\pi(T-K)}}\nonumber \\
     &\exp{\bigg(- \frac{T-K}{2H_2\sigma^2\log(K)}\bigg)}, \label{eqn:SR_Gaussian}\\
    p_{\rm G}^{\mathsf{SH}}(T) \leq &3 \log_2(K) \sqrt{\frac{2H_2\sigma^2\log_2(K)}{\pi T}}\nonumber \\
    &\exp{\bigg(-\frac{T}{8H_2\sigma^2\log_2(K)} \bigg)}.\label{eqn:SH_Gaussian}
\end{align}
\end{lemma}
We note the improvement in the leading term of the upper bound for the Gaussian reward setting. We also see that the effective hardness also introduces the variance parameter as a multiplicative term.
These results along with the same for our work are summarized in Table~\ref{table:1} and Table~\ref{table:2}.

\begin{table}
    \centering
    \begin{tabular}{|c||c|c|} 
     \hline
      %\multirow{}{}
      {Algorithm} 
      & \multicolumn{2}{c|}{Bounded Rewards}  \\ 
      \cline{2-3}
      & Leading term & Exponent term \\ 
     \hline\hline
      UE & $K-1$ & $2H_3$\\ 
     \hline
      SR &$K(K-1)/2$ & $H_2\log(K)$ \\
     \hline
      SH & $ 3 \log_2(K)$& $8H_2\log_2(K)$\\ 
     \hline
      { \Ac{RE} (ours)} & {$\log_2(K)$} & {$\frac{4\tilde{H}_4\log_2(K)}{\eta}(1 + \frac{\sqrt{K}}{3\sqrt{\tilde{H}_4}})$} \\ [1ex]
     \hline
    \end{tabular}
    \caption{Upper bounds for bounded rewards$^1$.}
    \label{table:1}
\end{table}

\blfootnote{$^1$The separability parameter $\eta$ is rigorously introduced and discussed in Section~III-B.}
\begin{table}
    \centering
    \begin{tabular}{|c||c|c|} 
     \hline
      %\multirow{}{}
      {Algorithm} 
      & \multicolumn{2}{c|}{Gaussian Rewards}  \\ [0.5ex]
      \cline{2-3}
      & Leading term & Exponent term \\ [0.5ex]
     \hline\hline
      UE & $(K-1)\sqrt{\frac{H_3\sigma^2}{\pi T}} $ & $4H_3\sigma^2$\\ [1ex]
     \hline
      SRs &$\frac{K(K-1)}{2} \sqrt{\frac{H_2\sigma^2\log(K)}{2\pi(T-K)}}$ & $2H_2\sigma^2\log(K)$ \\ [1ex]
     \hline
      SH & $3 \log_2(K) \sqrt{\frac{2H_2\sigma^2\log_2(K)}{\pi T}}$& $8H_2\sigma^2\log_2(K)$\\ [1ex]
     \hline
     { \ac{RE} (ours)} & {$\log_2(K)\sqrt{\frac{4\tilde{H}_4\sigma^2 \log_2(K)}{\pi \eta T}}$} & {$16\tilde{H}_4\sigma^2\log_2(K)/\eta$} \\ [1ex]
     \hline
    \end{tabular}
    \caption{Upper bounds for Gaussian rewards$^1$.}
    \label{table:2}
\end{table}

%\subsection{Objective}
Our objective is to improve upon the above algorithms in terms of the upper bounds by relaxing the condition that only one arm can be pulled per slot. Precisely, at each slot $t$, a subset of arms is sampled, i.e., $A_t \subseteq [K]$. %The reward received by the agent at time slot $t$ is $\sum_{i \in A_t} X_i(t)$, where $X_i(t) \sim \nu_i$.

\section{\ac{RE} Algorithm}
In this section, we present our algorithm based on grouping the arms based on Hamming codes and performing \acp{LRT} to detect the presence of the best arm. 
\begin{algorithm}[ht]
\caption{\ac{RE}}
\label{alg:rapid_exploration}
        {\bf Initialize:} Group the $K$ arms into $\log_2(K)$ groups of size $K/2$ based on Algorithm~\ref{alg:cap}.
         Estimate of mean of arm $i$ after $s$ rounds is $\hat{X}_{i,s}$, $1 \leq i \leq K$
        Estimate of mean of group $G$ after $s$ rounds is $\hat{\mu}_{G,s}$, $1 \leq G \leq \log_2(K)$.
        \begin{algorithmic}[1]
        \FOR{$t = 1 : \alpha T$}
            \STATE Pull each of the $K$ arms uniformly for $\alpha T/K$ times and observe the rewards.
        \ENDFOR
         \STATE Obtain estimates of means of each arm $\hat{X}_{i,\alpha T/K}$ for all arms.
        \STATE Obtain estimates of means of each group $\hat{\mu}_{G,\alpha T} = \frac{\sum_{i \in G}\hat{X}_{i,\alpha T/K}}{K/2}$ and calculate priors $\pi_{0,G}, \pi_{1,G}$ for each group for the presence of the best arm
        \FOR{$t = \alpha T : T$}
            \STATE Pull each of the $\log_2(K)$ groups uniformly and observe the sum of the rewards of the individual arms.
        \ENDFOR
        \STATE Perform \acp{LRT} for each group to test whether the best arm is present in that group.
        \STATE The best arm is the one that is common to all the groups in which it is detected.
    \end{algorithmic}
\end{algorithm}

\begin{algorithm}
\caption{Construct Groups}\label{alg:cap}
\begin{algorithmic}[1]
\STATE {\bf Input:} Arms and $K$.
\STATE {\bf Initialize:} $G_k = \{\}$, $\forall k = 1, 2, \ldots, \log_2 K$.
\FOR{$k = 1$ to $\log_2 K$}
    \FOR{$i = 0$ to $K - 1$}
        \IF{\texttt{dec2bin}$(i)$ \texttt{AND} \texttt{onehot}$(k) \neq \texttt{zeros}(1,n)$}
            \STATE $G_k = G_k \cup i$
        \ENDIF
    \ENDFOR
\ENDFOR
\STATE Return $\{G_k, 1\leq k \leq \log_2 K\}$
\end{algorithmic}
\end{algorithm}

\subsection{Arm Grouping and Decoding Strategy}
Our algorithm commences with first creating $\log K$ groups of arms denoted as $G_k, k \in [\log K]$. This is the same grouping strategy employed in \cite{ghatak2023fast} for change detection, whereas here, we employ it to employ \acp{LRT} discussed later. The $i$-th arm is added to a group $G_k$, if and only if the binary representation of $i-1$ has a "1" in the $k$-th binary place. In other words, arm $i$ is added to $G_k$ if
%\begin{align}
    $\texttt{bin2dec}\left(\texttt{dec2bin}(i-1) \; \texttt{ AND } \; \texttt{onehot}(k) \right) \neq 0,$
    %\nonumber 
%\end{align}
where $\texttt{bin2dec}()$ and $\texttt{dec2bin}()$ are respectively operators that convert binary numbers to decimals and decimal numbers to binary. Additionally, $\texttt{onehot}(k)$ is a binary number with all zeros except 1 at the $k$-th binary position. $\texttt{AND}$ is the bit-wise AND operator. The arm grouping strategy follows similarly to the parity bit generation using Hamming code, which belongs to the family of linear error-correcting codes~\cite{hamming1950error}. It is worth recalling that Hamming codes are perfect codes, that is, they achieve the highest possible rate for codes with their block length and minimum distance of three. Thus, the minimum number of parity bits needed to detect and correct one-bit errors is $\log_2 (K)$ for a $K-$ bit sequence. As a consequence of this, the minimum groups that are needed to detect a change in a single arm as well as identify the changed arm is $\log_2 (K)$. This is formally presented below.
\begin{remark}
    {Let $H \in \mathbb{F}_2^{m \times K}$ be the parity-check matrix of the binary Hamming code of length $K = 2^m - 1$. Then, for any best arm $a^\star \in \{1, \dots, K\}$, the group detection vector $y = H a^\star \pmod{2}$ uniquely identifies $a^\star$ with $m = \log_2(K+1)$ group tests. Moreover, this construction achieves the minimal number of tests necessary for unique identification in the binary group testing model.}    
\end{remark}
{ Each group consists of $\frac{K}{2}$ arms with optimal redundancy so as to enable the LLR framework to uniquely identify the best arm with the sequence of detections of presence and absence of the best arm in each of the groups.} Once the groups are formed, we define the mean of the group $G_k$ as the average of the means of the arms constituting the group, i.e., $\mu_{G_k} = \frac{2}{K}\sum_{i \in G_k}\mu_i$.

\subsection{Separability}
{A key assumption for the feasibility of the proposed algorithm is the {\it separability} of the arm groupings, i.e., the algorithm is able to differentiate a group which contains the best arm from a group which does not. In case the best arm is not present in a group, the mean of the group is upper bounded by the case that all the $K/2$ arms of that group have the same mean as the second best arm. Thus,
   $\mu_{G_k} \leq \mu_{[2]} = \mu_{[1]} - \Delta_{[2]}, \quad \forall k$, if $a^* \notin G_k.$ Similarly, in case the best arm is present in a group, the lower bound of the mean of this group is obtained by assuming that the remaining arms of that group consist of the arms with the lowest means as
   \begin{equation}
    \mu_{G_k} \geq \frac{\mu_{[1]} + (K/2 - 1)\mu_{[K]}}{K/2},  \quad \forall k,  \nonumber
\end{equation} if $a^* \in G_k$. For these two quantities to be sufficiently spaced, we impose the following,\\
\begin{assumption} [separability for worst-case prior] To guarantee that a group containing the best arm is distinguishable from a group that does not contain the best arm under our grouping construction we impose the separability condition 
\begin{equation}\label{eqn:assumption1}
   \Delta_{[2]} - (1-2/K)\Delta_{[K]}\geq \frac{\sqrt{\eta}(\Delta_{[1]} + \Delta_{[K]})}{K}= \frac{1}{K}\sqrt{\frac{\eta}{H_4}}
\end{equation}
where $\eta \leq 1$ is a fixed constant. We emphasize that this condition is a sufficient separability requirement used to obtain the clean bounds in Sections IV and V and that it constrains the relative sizes of the largest and second-smallest suboptimality gaps.
\end{assumption} 
\begin{remark}
    The separability condition ensures that even in the worst case, where all arms in a group without the best arm are as large as \(\mu_{[2]}\), the group mean remains \textit{sufficiently} below the smallest possible group mean that contains the best arm, and therefore the LRT can discriminate the two hypotheses.
\end{remark}
\begin{remark}
   By ``large-gap'' regime we mean that the gap between the best arm and its closest competitor \(\Delta_{[1]}=\Delta_{\min}\) is large relative to observation noise; this is different from requiring large dispersion among the suboptimal arms.
\end{remark}
The separability condition in Assumption \ref{eqn:assumption1} is conservative and sufficient for our analysis.}
{
\begin{remark}
    Assumption 1 is necessary to ensure that multiple sub-optimal arms cannot be grouped together to drive the group mean close to the optimal arm's mean. In case this assumption does not hold, the likelihood ratio test based detection of the presence of the best arm in a group may fail with an arbitrarily high probability.
\end{remark}
\begin{remark}
    It is important to note that the baseline algorithms for single play \ac{BAI} such as \ac{SR}~\cite{audibert2010best} and \ac{SH}~\cite{karnin2013almost} indeed perform more efficiently in terms of the probability of error if Assumption 1 holds - a fact that is characterized by the resulting hardness of the problem under Assumption 1. However, in case the number of arms $K$ is large, the large-gap condition of Assumption 1 does not improve the scaling of the performance with respect to $K$. On the contrary, given Assumption 1 holds, the algorithm proposed in our work achieves a better performance than baseline algorithms. This due to the fact that a large-gap along with multi-play implies that the player may be able to observe multiple instances of the best arm (as a part of multiple groups) against each instance for baseline algorithms.
\end{remark}
}
Before the formation of the groups, { in case the player does not have information about the minimum sub-optimality gap, $\mu_{[1]}$} we perform an {initial exploration phase of length $\alpha T$ where {\it each arm} is probed for $\alpha T/K$ slots, $0 \leq \alpha < 1$.} This is for constructing initial estimates of the means of the arms, { for creating an estimate $\hat{\mu}_{[1]}$ of the minimum sub-optimality gap,} and to engineer the prior probabilities of the hypothesis testing. Once the arms are grouped together, the groups are pulled followed by the composite hypothesis testing as discussed below. { Thus, the overall algorithm has two phases -- one optional initial exploration phase followed by the RE phase.} { An example of such explore-the-commit procedure is presented in~\cite{nie2022explore}.}

\subsection{Hypothesis Test}
{In what follows, we drop the subscript $k$ from $G_k$ since the analysis holds equivalently for all the groups. Let $H_{1,G}$ represent the hypothesis that the best arm is present in group $G$, and $H_{0,G}$ represent the hypothesis that the best arm is absent in group $G$, i.e.,
\begin{align}
    &H_{1,G} := \{\mu_{[1]} \in G\}; \qquad H_{0,G} := \mu_{[1]} \notin G   \nonumber.
\end{align}
Since as per Assumption \ref{eqn:assumption1}, the presence of the best arm in a group enhances the mean of the corresponding group, let us denote by $\Lambda_1$ the range in which the mean of a group containing the best arm lies. Similarly, we denote by $\Lambda_0$, the range in which the mean of a group lies in case it does not contain the best arm. Due to our assumptions, we have $\Lambda_1 := [\mu_{[1]}- (1-2/K)\Delta_{[K]}, \mu_{[1]} - (1-2/K)\Delta_{[2]}]$, and $\Lambda_0 = [\mu_{[1]}-\Delta_{[K]}, \mu_{[1]} - \Delta_{[2]}]$. Thus, the hypothesis groups have the following equivalences.
\begin{align}
    &H_{1,G} := \{\mu_{[1]} \in G\}  \nonumber \\ \nonumber
            &\equiv {\mu}_G \in \Lambda_1 = [\mu_{[1]}- (1-2/K)\Delta_{[K]}, \mu_{[1]} - (1-2/K)\Delta_{[2]}],
\\
    &H_{0,G} := \mu_{[1]} \notin G   \nonumber
            \equiv {\mu}_G \in \Lambda_0 = [\mu_{[1]}-\Delta_{[K]}, \mu_{[1]} - \Delta_{[2]}].
\end{align}
Then, based on the observed rewards for $t$ pulls of the group $G$, let us represent the likelihood under both the hypothesis with the help of corresponding joint probability distributions $p(R_{G,1:(1-\alpha)T/\log_2(K)} | \mu_{[1]} \notin G)$ and $p(R_{G,1:(1-\alpha)T/\log_2(K)} | \mu_{[1]} \in G)$, respectively for $H_{0,G}$ and $H_{1,G}$, respectively. The exact mathematical form of the likelihood depends on the underlying reward distributions and is further investigated in the next two sections, specifically for Gaussian and Bounded rewards. The general rule for the likelihood ratio test for composite hypotheses for identifying the presence/absence of the best arm for the worst case prior in the above problem is \cite{DnE_book}
\begin{align}
&\delta_G = \nonumber \\
   & \begin{cases}
        0; &\text{if}\quad  \mathcal{L}  := \frac{\min_{\mu_G \in \Lambda_1}p(R_{G, 1:(1-\alpha)T/\log_2(K)} | \mu_{[1]} \in G)}{\max_{\mu_G \in \Lambda_0}p(R_{G,(1:1-\alpha)T/\log_2(K)} | \mu_{[1]} \notin G)} \leq \frac{\pi_{0,G}}{\pi_{1,G}}; \nonumber \\
        1; & \text{otherwise},
    \end{cases}
\end{align}
where $\delta_G$ is the indicator function determining if the best arm is present in the group $G$. In the above, we have considered uniform assignment costs $(C_{10} = C_{01} = 1, C_{11} = C_{00} = 0)$, i.e., a cost 0 in case of a correct assignment, e.g., $\delta_G = 0$ under $H_{0,G}$, and a cost 1 (or equivalently any constant) in case of an incorrect assignment, e.g., $\delta_G = 1$ under $H_{0,G}$.
Furthermore, $R_{G,1:t}$ represents the rewards from group $G$ after $t$ pulls. For an exploration fraction $\alpha$, in our framework, $\alpha T$ pulls are reserved for initial exploration (as elaborated in the next section on engineering the priors. Accordingly, $(1-\alpha)T$ pulls remain for detecting the presence of the best arm in the groups. The priors $\pi_{0,G}, \pi_{1,G}$ represent our beliefs on whether the best arm is present or absent in group $G$ based on our initial exploration for $\alpha T$ rounds, which is elaborated in the next sub-section.}

\subsection{Initial Exploration and Engineering the Priors}
In order to employ the composite hypothesis test described above, we require the priors $\pi_{0,G}, \pi_{1,G}$ which represent the probabilities that the best arm is absent or present in group $G$. Since this is not present with the agent, it can be engineered with the information gained during the initial uniform exploration stage for $\alpha T$ rounds. Recall that during the initial exploration stage, each arm is pulled $\alpha T/K$ times and we obtain estimates of means of all arms $\hat{X}_{i,\alpha T/K}$ $(1\leq i\leq K)$. Using these estimates, we then obtain the mean estimates of each of the groups $\hat{\mu}_{G, \alpha T}$ $(1\leq G\leq \log_2(K))$ as
\begin{align}
        \hat{\mu}_{G, \alpha T} = \frac{\sum_{i \in G} \hat{X}_{i,\alpha T/K}}{K/2}.
\end{align}
Let the means of the group under hypotheses $H_{0,G}$ and $H_{1,G}$ be $\mu_L$ and $\mu_H$, respectively. Note that $\mu_H$ and $\mu_L$ are random variables whose distribution depends on how the distribution of each arm at each index is decided. Their expected values are given by $\bE_{\nu}[\mu_G | \mu_{[1]} \in G], \bE_{\nu}[\mu_G | \mu_{[1]} \notin G]$. Since the priors sum up to 1, we can view the pair as the parameters of a Bernoulli distribution. In other words, the event that the best arm belongs to this group or not is a Bernoulli random variable with parameters given by the above means. We substitute $\bE[\mu_H] = \mu_{[1]} - (1-2/K)(\Delta_{\min} + \Delta_{\max})/2$ and $\bE[\mu_L] = \mu_{[1]} - (\Delta_{\min} + \Delta_{\max})/2$. This is because if the arms are arranged randomly among the $K$ positions, and the means of the suboptimal arms are chosen uniformly at random from $[\mu_{[1]} - \Delta_{\max}, \mu_{[1]} - \Delta_{\min}]$, we can obtain the substituted values of $\bE[\mu_H]$ and $\bE[\mu_L]$ %Recall that the Bernoulli distribution is a member of the exponential family with the parameter being the sigmoid of the natural parameter~\cite{GLM}.
Thus, introducing sigmoids into the formulation of the priors is a natural choice~\cite{GLM}. However, a key issue is the asymmetric nature of the domain as the composite hypotheses $H_{1,G}$ and $H_{0,G}$ have supports of different sizes. To handle this asymmetry, we will construct two sigmoids, and normalize them appropriately. We now define $\sigma_{in}(x)$ and $\sigma_{out}(x)$ which represent the priors that the best arm is present or absent in group $G$:
\begin{align}
    \sigma_{in}(x) &= \dfrac{1}{1+\exp(-(x-\bE[\mu_H])/|\Lambda_1|)}, \nonumber \\
    \sigma_{out}(x) &= \dfrac{1}{1 + \exp((x-\bE[\mu_L])/|\Lambda_0|)}, \nonumber
\end{align}
where $|\Lambda_0|, |\Lambda_1|$ represent the size of the intervals corresponding to the hypotheses $H_{0,G}, H_{1,G}$. We further normalize them to obtain the engineered $\pi_{0,G}, \pi_{1,G}$ as
\begin{equation}
    \pi_{0,G} = \frac{\sigma_{out}(\hat{\mu}_G)}{\sigma_{out}(\hat{\mu}_G) + \sigma_{in}(\hat{\mu}_G)}, \hspace{0.5cm}
    \pi_{1,G} = \frac{\sigma_{in}(\hat{\mu}_G)}{\sigma_{out}(\hat{\mu}_G) + \sigma_{in}(\hat{\mu}_G)}.
\end{equation}
This is a good choice for priors as $\sigma_{in}(\cdot)$ is an increasing function of $\hat{\mu}_G$ and $\sigma_{out}(\cdot)$ is a decreasing function of $\hat{\mu}_G$. Also, the arguments are normalized such that the majority of the rise/fall of the sigmoid occurs in their respective hypothesis intervals, i.e., $\Lambda_1$ and $\Lambda_0$, respectively.
    { 
    \begin{remark}
        Note that for the application of RE, the player either needs to have prior knowledge of $\pi_{0,G}$ and $\pi_{1,G}$, e.g., as a side information or needs to estimate it using the framework described in this sub-section. In this regard, we note that during the initial exploration phase, the mean rewards for all the arms are estimated and accordingly, the player will have an estimate of $\hat{\mu}_{[1]}$ before initiating the RE phase. With appropriate design of the length of the initial exploration phase, the probability that the estimate deviates from the true value can be limited, e.g., with an $\mathcal{O}(\log(n))$ long initial exploration phase, the probability that the estimate of $\mu_{[1]}$ is incorrect is exponentially bounded. Nevertheless, in this work, we focus on applications where the player will have a priori knowledge of $\mu_{[1]}$. As we will see in the case studies later, in the jammer design problem, the jammer knows a-priori the received signal statistics when the correct waveform is selected, albeit not knowing the identity of the correct waveform. Similarly, for the radar detection problem, the task of the detector is to identify the active channel while having prior information of the radar signature.
    \end{remark}
    }
\subsection{Recommendation of the Best Arm}
The final recommendation of the best arm follows the results of the hypothesis tests of the individual groups. We employ a Hamming decoding method, wherein, the best arm is the one that belongs to the all groups in which in which it is detected. Thanks to Hamming codes, the binary output of the presence or the absence of the best arm in each group uniquely determines the best arm. An error in determining the best arm occurs when at least one of the group \acp{LRT} fail. In the next section, we analyze the probability of error of the proposed algorithm. We consider 2 cases, one for Gaussian reward distribution and the other for uniform reward distribution. The total probability of error is given by
\begin{align}
    p_{(\cdot)}^{\rm RE} \triangleq 1 - \prod_{i = 1}^{\log_2 K} (1-\cP_{e,G_i}), \nonumber
\end{align}
where $\cP_{e,G_i} = \mathbb{P}\left(\delta_{G_i} , H_{1,G_i}\right) + \mathbb{P}\left(\delta^c_{G_i}, H_{0,G_i}\right)$. Assume that group $G$ has the highest probability of error, this allows us to bound $p^{\rm RE}_{(\cdot)}$ as
%\begin{align}
    $p^{\rm RE}_{(\cdot)} \leq 1 - (1-\cP_{e,G})^{\log_2(K)} \leq \log_2(K) \cP_{e,G}$. %\nonumber 
%\end{align}

\section{Probability of Error with Gaussian Rewards With Perfect Knowledge of $\mu_{[1]}$}
{The distribution of the reward of a group, conditioned on the presence or absence of the best arm in the group are
    \begin{align}
        R_{G,t}|\mu_{[1]} \in G &\sim \cN(\mu_H, \sigma^2/(K/2)), \; \text{and} \nonumber \\
         R_{G,t}|\mu_{[1]} \notin G &\sim \cN(\mu_L, \sigma^2/(K/2)). \nonumber
    \end{align}
Let $\overline{R}_{G,t}$ be the mean reward of the group $G$ after $t$ pulls. Since each group is pulled $(1-\alpha)T/\log_2(K)$ times, the conditional distributions of the group means are given by
\begin{align}
    \overline{R}_{G, (1-\alpha)T/\log_2(K)}|\mu_{[1]} &\in G \sim \cN\bigg(\mu_H, \frac{2\sigma^2\log_2(K)}{(1-\alpha)KT}\bigg) \nonumber
\\
    \overline{R}_{G, (1-\alpha)T/\log_2(K)}|\mu_{[1]} &\notin G \sim \cN\bigg(\mu_L, \frac{2\sigma^2\log_2(K)}{(1-\alpha)KT}\bigg) \nonumber
\end{align}
The composite hypotheses are
\begin{align}
H_{1,G}:~ \mu_G \in \Lambda_1=[\mu_H^{\min}, \mu_H^{\max}], 
\\
H_{0,G}:~ \mu_G \in \Lambda_0 = [\mu_L^{\min},\mu_L^{\max}].
\end{align}
and the exact distributions of $\mu_H$ and $\mu_L$ follow the Irwin-Hall distribution (sums of uniform random variables) as the suboptimality gaps are sampled uniformly at random from the interval $[\Delta_{[1]}, \Delta_{[K]}]$. Since the Gaussian family $\{ \mathcal{N}(\mu,\sigma_G^2)\}$ has a monotone
likelihood ratio in $\mu$, classical results (Chernoff, 1954; Lehmann--Romano)
guarantee that the pair
\[
\mu_L^\star = \sup \Lambda_0 = \mu_L^{\max},
\qquad
\mu_H^\star = \inf \Lambda_1 = \mu_H^{\min}
\]
\emph{maximizes} the probability of error among all distributions in the composite
hypotheses. Therefore designing the likelihood-ratio test for the simple
hypotheses
\[
\cN\bigg(\mu_L^\star, \frac{2\sigma^2\log_2(K)}{(1-\alpha)KT}\bigg)
~~\text{vs.}~~ 
\cN\bigg(\mu_H^\star, \frac{2\sigma^2\log_2(K)}{(1-\alpha)KT}\bigg)
\]
automatically provides a uniformly valid test for the entire composite hypotheses.
Thus, the LRT is
    \begin{align}
       \mathcal{L} &= \frac{\min_{\mu_H \in \Lambda_1}p(R_{G, 1:(1-\alpha)T/\log_2(K)} | \mu_{[1]} \in G)}{\max_{\mu_L \in \Lambda_0}p(R_{G, 1:(1-\alpha)T/\log_2(K)} | \mu_{[1]} \notin G)}   \nonumber \\
       &=
        \frac{\exp\bigg(-\frac{(1-\alpha)TK(\overline{R}_{G,(1-\alpha)T/\log_2(K)} - \mu_H^\star)^2}{4 \sigma^2 \log_2(K)}\bigg)}{\exp\bigg(-\frac{(1-\alpha)TK(\overline{R}_{G,(1-\alpha)T/\log_2(K)} - \mu_L^\star)^2}{4 \sigma^2 \log_2(K)}\bigg)}, 
    \end{align}
    and the final decision rule now simplifies to 
     \begin{align}
        \overline{R}_{G,(1-\alpha)T/\log_2(K)} &\lesseqgtr \frac{\mu_H^\star + \mu_L^\star}{2} + \frac{2\sigma^2\ln(\pi_{0,G}/\pi_{1,G})\log_2(K)}{(1-\alpha)TK(\mu_H^\star-\mu_L^\star)} \nonumber \\
        &\triangleq \tau_G \nonumber
    \end{align}

    \begin{remark}
We assume \(K = 2^m\) for theoretical analysis. If \(K\) is not a power of two, we augment the arm set to \(\widetilde{K} = 2^{\lceil \log_2 K \rceil}\) by adding \(\widetilde{K}-K\) dummy arms with mean \(\mu_{\mathrm{dummy}} \ll \mu[K]\). This ensures each group contains exactly \(\widetilde{K}/2\) arms. Under this padded representation, the variance of the group-level reward becomes $\mathrm{Var}(R_{G,t}) = \frac{2\sigma^2}{\widetilde{K}},$ and the likelihood ratio threshold is updated to
\[
   \tau_G = \frac{\mu_H^\star + \mu_L^\star}{2} + \frac{2\sigma^2\log_2(\widetilde{K})\ln(\pi_{0,G}/\pi_{1,G})}{(1-\alpha)\widetilde{K}T(\mu_H^\star - \mu_L^\star)}.
\]
The dummy arms do not affect the identity of the optimal arm and only enable consistent group sizing required by the binary construction.
    \end{remark}
    
    Now, the algorithm can make a mistake when either a \textit{false alarm} (Type-I error) or a \textit{missed detection} (Type-II error) occurs. A \textit{false alarm} occurs when the ground truth is that the best arm does not belong to the group but the LRT outputs $H_1$, and a \textit{missed detection} occurs when the ground truth is that the best arm is present in the group but the LRT outputs $H_0$. 
    Thus, the probability of error of a single hypothesis test for group $G$ can be written as
    \begin{align}
        &\cP_{e,G} = \mathbb{P}\left(\delta_G, H_{1,G}\right) + \mathbb{P}\left(\delta^c_G , H_{0,G}\right) \nonumber \\
        &= \Tilde{\pi}_{0,G}\bP_{\nu}\left(\delta^c_G  \Large| H_{0,G}\right) + \Tilde{\pi}_{1,G}\bP_{\nu}\left(\delta_G  \Large| H_{1,G}\right). \label{eqn:grp_error_prob}
    \end{align}
Note that $\Tilde{\pi}_{0,G}$ and $\Tilde{\pi}_{1,G}$ are the actual probabilities of the best arm being absent and present in group $G$ and not the estimates $\pi_{0,G}, \pi_{1,G}$ which have been obtained. The \textit{false alarm} and \textit{missed detection} probabilities can be expressed in terms of the tail probability of Gaussian random variables as,
    \begin{align}
        \mathcal{P}_{\rm F} & = \bP_{\nu}\left(\delta_G| H_{0,G}\right) \nonumber \\
        &= \mathbb{P}\bigg(\overline{R}_G > \tau_G| \overline{R}_G \sim \mathcal{N}\bigg(\mu_L^\star, \frac{2\sigma^2\log_2(K)}{(1-\alpha)TK}\bigg)\bigg) \nonumber \\
        &= \mathbb{P}\bigg(\mathcal{N}(0,1) > (\tau_G - \mu_L^\star)\sqrt{\frac{(1-\alpha)TK}{2\sigma^2 \log_2(K)}}\bigg) \nonumber \\
        &= Q\bigg((\tau_G - \mu_L^\star)\sqrt{\frac{(1-\alpha)TK}{2\sigma^2 \log_2(K)}}\bigg).
    \end{align}
    \begin{align}
        \mathcal{P}_{\rm M} = &\bP_{\nu}\left(\delta^c_G | H_{1,G}\right) \nonumber \\
        &= \mathbb{P}\bigg(\overline{R}_G < \tau_G| \overline{R}_G \sim \mathcal{N}\bigg(\mu_H^\star, \frac{2\sigma^2\log_2(K)}{(1-\alpha)TK}\bigg)\bigg) \nonumber \\
        &= \mathbb{P}\bigg(\mathcal{N}(0,1) < (\tau_G - \mu_H^\star)\sqrt{\frac{(1-\alpha)TK}{2\sigma^2 \log_2(K)}}\bigg) \nonumber \\
        &= Q\bigg((\mu_H^\star - \tau_G)\sqrt{\frac{(1-\alpha)TK}{2\sigma^2 \log_2(K)}}\bigg),
    \end{align}
    where 
    \begin{equation}
        Q(x) = \frac{1}{\sqrt{2\pi}}\int_{x}^{\infty} e^{-t^2/2} dt.
    \end{equation}
    We now bound the probability of error with the maximum of the \textit{false alarm} and \textit{missed detection} probabilities. Define $\pi_{high,G} = \max\{\pi_{0,G}, \pi_{1,G}\}$ amd $\pi_{low,G} = \min\{\pi_{0,G}, \pi_{1,G}\}$. Thus, 
    \begin{align}
        \cP_{e,G} &\leq \max\{\mathcal{P}_{\rm F} , \mathcal{P}_{\rm M}\} \nonumber \leq  Q\bigg(\frac{\mu_H^\star - \mu_L^\star}{2} \sqrt{\frac{(1-\alpha)TK}{2\sigma^2\log_2(K)}}- \nonumber \\
        &\sqrt{\frac{2\sigma^2\log_2(K)}{(1-\alpha)TK}}\frac{\ln(\pi_{high,G}/\pi_{low,G})}{(\mu_H^\star-\mu_L^\star)}\bigg). \label{eqn:max FA,MD}
    \end{align}
    From \cite{Q_function_bound}, we have the following bounds on $Q(\cdot)$
    \begin{equation}\label{eqn:Qbound}
        \frac{x}{(1+x^2)\sqrt{2\pi}} e^{-x^2/2} < Q(x) < \frac{1}{x\sqrt{2\pi}}e^{-x^2/2}.
    \end{equation}
    Therefore, by applying (\ref{eqn:Qbound}) and Assumption \ref{eqn:assumption1}, we get 
\begin{theorem}\label{thm:1}
{For Gaussian rewards with constant variance $\sigma^2$
, and $\alpha = 0$, the error probability of RE is bounded as
    \begin{equation}\label{eqn:RE_Gaussian}
         p^{\rm RE}_{G} \leq \sqrt{\frac{4H_4\sigma^2 K\log^3_2(K)}{\pi \eta T}}\exp\bigg(\frac{-\eta T}{16H_4\sigma^2K\log_2(K)}\bigg).
    \end{equation}}
\end{theorem}
Note that for a bandit model $\nu$ where all the non-optimal arms have the same mean $(\Delta_{[2]} = \Delta_{[K]}$, we have $H_1 = H_2 = H_3 = 4KH_4$, and equality in Assumption \ref{eqn:assumption1} holds with $\eta = 1$. Thus, plugging this value of $H_4$ into (\ref{eqn:RE_Gaussian}) gives a better result than SR and SH. Thus, for a problem instance with a large number of arms, the \ac{RE} algorithm gives us a lower probability of error than the algorithms designed for non-combinatorial bandits. It also performs better than Combinatorial Successive Accept Reject (CSAR) algorithm of \cite{CPE}. The CSAR algorithm also requires access to a \textit{constrained oracle} to compute the optimal set to sample in each phase of the algorithm.
\begin{corollary}
\label{cor:single-gap}
Suppose that all sub-optimal arms share the same sub-optimality gap, i.e.,
\begin{align}
\Delta_{[2]} = \Delta_{[3]} = \cdots = \Delta_{[K]} = \Delta, \nonumber \\
H_1 = H_2 = H_3 = 4 K H_4 = K/\Delta^2,
\end{align}
for some $\Delta>0$. Then the distributions of the group means under the two hypotheses degenerate to point masses, and the likelihood ratio test in Theorems~\ref{thm:1} and~\ref{thm:2} reduces to a simple hypothesis test with
\[
\mu_H^\star = \mu_{[1]} - \Bigl(1 - \frac{2}{K}\Bigr)\Delta,
\qquad
\mu_L^\star = \mu_{[1]} - \Delta.
\]
In this case, the separability condition in Assumption~1 holds with equality for $\eta = 1$, which is the \emph{largest possible} value of~$\eta$ and therefore corresponds to the strongest achievable separation between the two hypotheses. As the error bounds in Theorems~\ref{thm:1} and~\ref{thm:2} improve monotonically with~$\eta$, this single-gap regime yields the smallest attainable probability of error for the RE algorithm. Thus, plugging in this value of $H_4$ and $\eta$ in this best-case scenario, RE achieves a strictly tighter exponential rate than both Successive Rejects (SR) and Successive Halving (SH) which are algorithms designed for non-combinatorial bandits.  It also performs better than Combinatorial Successive Accept Reject (CSAR) algorithm of \cite{CPE}. The CSAR algorithm also requires access to a \textit{constrained oracle} to compute the optimal set to sample in each phase of the algorithm.
\end{corollary}
    \begin{remark}
Beyond the single-gap setting, the group means under $H_{0,G}$ and $H_{1,G}$ can be expressed as averages of (transformed) i.i.d.\ sub-optimality gaps. Under the generative model in Section~III-D, the gaps $\{\Delta_a : a \neq a^\star\}$ are sampled i.i.d.\ from a uniform distribution on $[\Delta_{\min}, \Delta_{\max}]$. Consequently, the group means $\mu_L$ (best arm absent) and $\mu_H$ (best arm present) are sums of independent uniform random variables and thus follow Irwin--Hall distributions with parameters
\begin{align*}
 \mathbb{E}[\mu_L]
  = \mu_{[1]} - \frac{\Delta_{\min}+\Delta_{\max}}{2}, \\
  \operatorname{Var}(\mu_L)
  = \frac{(\Delta_{\max} - \Delta_{\min})^2}{6K}\\
  \mathbb{E}[\mu_H]
  = \mu_{[1]} - \left(1 - \frac{2}{K}\right)\frac{\Delta_{\min}+\Delta_{\max}}{2}, \\
  \operatorname{Var}(\mu_L)
  = \frac{(\Delta_{\max} - \Delta_{\min})^2}{6K}\left(1 - \frac{2}{K}\right)
\end{align*}
 By the classical Central Limit Theorem, we therefore obtain
\[
\mu_L \xrightarrow[]{d} \mathcal{N}(\mathbb{E}[\mu_L], \operatorname{Var}(\mu_L)),
\qquad
\mu_H \xrightarrow[]{d} \mathcal{N}(\mathbb{E}[\mu_H], \operatorname{Var}(\mu_H)),
\]
with $\operatorname{Var}(\mu_H), \operatorname{Var}(\mu_L) \to 0$ as $K \to \infty$. In other words, for large $K$, both $\mu_H$ and $\mu_L$ are tightly concentrated around their respective means.
Figure \ref{fig:distribution} illustrates this effect: even for $K=16$, the empirical distributions of $\mu_H$ and $\mu_L$ align closely with their Gaussian CLT approximations. This empirically supports that, in high-arm regimes, replacing the random group means by their typical values is a good approximation.
Finally, combining this concentration with Assumption~1 yields the following asymptotic picture as $K \to \infty$:
\begin{itemize}
    \item The spread between the smallest and largest sub-optimality gaps, captured by $\Delta_{[2]}$ and $\Delta_{[K]}$, shrinks relative to the scale set by the hardness parameter $H_4$.
    \item The random fluctuations in $\mu_H$ and $\mu_L$ vanish as their variance scales as $\mathcal{O}(1/K)$, so the group means behave as if they were fixed at their expectations.
\end{itemize}
Thus, in the regime with $K \to \infty$, the strengthened separability condition becomes effectively tight: the random fluctuations in $\mu_H$ and $\mu_L$ vanish, and the least favorable gap 
$\mu_H^\star - \mu_L^\star$ from Assumption \ref{eqn:assumption1} achives equality with $\eta \to 1$.
\end{remark}}
\begin{figure}[!h]
    \centering
    \includegraphics[width=0.75\linewidth]{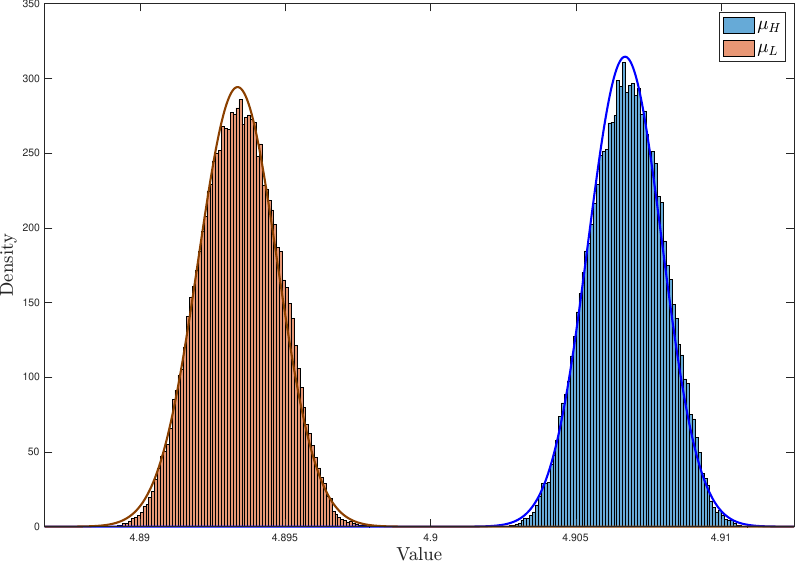}
    \caption{Empirical histograms of $\mu_H$ and $\mu_L$ for $K = 16$ along with Gaussian distributions (solid lines) obtained via the CLT.}
    \label{fig:distribution}
\end{figure}

\section{Probability of Error with Bounded Rewards With Perfect Knowledge of $\mu_{[1]}$}
{Unlike the previous section, where we used the $Q(\cdot)$ function to bound the error probability, we make use of Bernstein's inequality to bound the probability of error~\cite{giroux1979bernstein}. The group reward observations are i.i.d and the variance is decreased by a factor of $K/2$, which is the group size. Thus for $\alpha = 0$, the probability of error of the RE algorithm can be bounded as,
\begin{align}
    &\bP_{\nu}\left(\delta_G  | H_{0,G}\right) = \bP\bigg(\overline{R}_{G, T/\log_2(K)} > \frac{\mu_H^\star + \mu_L^\star}{2} | \mu_{[1]} \notin G\bigg) \nonumber \\
    &= \bP\bigg(\overline{R}_{G, T/\log_2(K)} - \mu_L^\star > \frac{\mu_H^\star - \mu_L^\star}{2} | \mu_{[1]} \notin G\bigg) \nonumber\\
    &\leq \exp\bigg(-\frac{T(\mu_H^\star - \mu_L^\star)^2}{8\log_2(K)(\frac{\sigma_{max}^2}{K/2} + \frac{\mu_H - \mu_L}{6})}\bigg) \nonumber\\
    &\leq \exp\bigg(-\frac{\eta T}{8K^2\log_2(K)H_4(\frac{2\sigma_{max}^2}{K} + \frac{1}{6K\sqrt{H_4}})}\bigg) \nonumber\\
    &\leq \exp\bigg(-\frac{\eta T}{8K\log_2(K)H_4(\frac{1}{2} + \frac{1}{6\sqrt{H_4}})}\bigg).
\end{align}
\begin{align}
    &\bP_{\nu}\left(\delta^c_G | H_{1,G}\right) = \bP\bigg(\overline{R}_{G, T/\log_2(K)} < \frac{\mu_H^\star + \mu_L^\star}{2} | \mu_{[1]} \in G\bigg) \nonumber\\
    &= \bP\bigg(\overline{R}_{G, T/\log_2(K)} - \mu_H^\star < -\frac{\mu_H^\star - \mu_L^\star}{2} | \mu_{[1]} \in G\bigg) \nonumber\\
    &\leq \exp\bigg(-\frac{T(\mu_H^\star - \mu_L^\star)^2}{8\log_2(K)(\frac{\sigma_{max}^2}{K/2} + \frac{\mu_H - \mu_L}{6})}\bigg) \nonumber\\
    &\leq \exp\bigg(-\frac{\eta T}{8K^2\log_2(K)H_4(\frac{2\sigma_{max}^2}{K} + \frac{1}{6K\sqrt{H_4}})}\bigg) \nonumber\\
    &\leq \exp\bigg(-\frac{\eta T}{8K\log_2(K)H_4(\frac{1}{2} + \frac{1}{6\sqrt{H_4}})}\bigg).
    \end{align}
    Similar to (\ref{eqn:max FA,MD}), we get
    \begin{equation}
        \cP_{e,G} \leq \exp\bigg(-\frac{\eta T}{8K\log_2(K)H_4(\frac{1}{2} + \frac{1}{6\sqrt{H_4}})}\bigg).
    \end{equation}
 \begin{theorem}\label{thm:2}
For bounded rewards in $[0,1]$ and $\alpha = 0$, the error probability of RE is bounded as
    \begin{equation}\label{eqn:RE_Bounded}
        p^{\rm RE}_B \leq \log_2(K)\exp\bigg(-\frac{\eta T}{8H_4K\log_2(K)(\frac{1}{2} + \frac{1}{6\sqrt{H_4}})}\bigg)
    \end{equation}
\end{theorem}
Thus, there is a clear advantage in the derived bound in the leading term. However, based on the value of the hardness parameter, experimentally often the \ac{RE} algorithm under performs. It is important to note that even though he upper bound of the probability of error provides a key design guideline in terms of the expected performance of an algorithm, often practical settings reveal its essential limitations. This can either be due to low stability in the performance (e.g., high variance) which is not captured in the upper bound of the probability of error, or due to the fact that the assumptions may not hold in real-world data. In order to rigorously study this we perform two case-studies which helps us highlight the conditions under which the empirical performance of \ac{RE} corroborates the theoretical guarantees. Furthermore, it helps highlight the detrimental impact on the performance when the key assumption of separability does not hold.}

\section{Probability of Error Due to the Initial Exploration Phase}
{ Recall that in our framework, in the initial exploration phase, each arm is sampled $m = \alpha T / K$ times. Let us define a {\it bad} event as $E^{(m)}_\epsilon := \left\{ \exists i \in [K] : |\hat{\mu}_i - \mu_i| > \epsilon \right\},$ which denotes the failure of estimation for at least one arm. Note that from the union bound of errors, the overall error probability can be bounded as follows.
\begin{align}
   \mathbb{P}(\text{RE fails}) 
\leq 
\underbrace{\mathbb{P}(E_\epsilon^{(m)})}_{\mathbb{P}(\text{Bad event})} 
+ 
\underbrace{{p}^{\rm RE}_G |\bar{E}_\epsilon^{(m)} }_{\mathbb{P}(\text{RE phase fails} \mid \text{good event})} 
\end{align}
In the previous sections, we have characterized ${p}^{\rm RE}_G |\bar{E}_\epsilon^{(m)}$. Here, we integrate it with $\mathbb{P}(E_\epsilon^{(m)})$. First, leveraging the Gaussian distribution, we have
\begin{align}
 &\bP(E^m_{\epsilon}) 
    = \bP\big(\bigcup_{i=1}^K |\hat{\mu_i} - \mu_i| > \epsilon \big) 
    \leq \sum_{i=1}^K  \bP\big(|\hat{\mu_i} - \mu_i| > \epsilon\big) \nonumber \\
    &=   \sum_{i=1}^K \bP\Bigg(\left|\frac{\sum_{t=1}^m X_{i,t}}{m} - \mu_i\right| > \epsilon \Bigg) =  \sum_{i=1}^K \bP(|\mathcal{N}(0, \sigma^2/m)| > \epsilon) \nonumber \\
    &=  \sum_{i=1}^K  \bP\bigg(|\mathcal{N}(0,1)| > \frac{\epsilon\sqrt{m}}{\sigma}\bigg) = 2KQ\bigg(\frac{\epsilon\sqrt{m}}{\sigma}\bigg) := \delta_1
\end{align}
Thus, with probability at least $1 - \delta_1$,  $ |\hat{\mu}_{[1]} - \mu_{[1]}| \leq \epsilon$,  $|\hat{\Delta}_{\min} - \Delta_{\min}| \leq \epsilon,$ and $|\hat{\Delta}_{\max} - \Delta_{\max}| \leq \epsilon$. Now recall that we have built the LRT threshold to be used in the RE phase as follows.
\begin{align}
    \tau_G = \frac{\mu_H + \mu_L}{2} + \frac{2\sigma^2 \ln\left( \frac{\pi_{0,G}}{\pi_{1,G}} \right) \log_2 K}{(1 - \alpha)TK(\mu_H - \mu_L)}\nonumber 
\end{align}
Since we use estimated values \( \hat{\mu}_{[1]}, \hat{\Delta}_{\min}, \hat{\Delta}_{\max} \), and estimated priors \( \hat{\pi}_{0,G}, \hat{\pi}_{1,G} \), the actual threshold implemented is
\begin{align}
    \hat{\tau}_G = \frac{\hat{\mu}_H + \hat{\mu}_L}{2} + \frac{2\sigma^2 \ln\left( \frac{\hat{\pi}_{0,G}}{\hat{\pi}_{1,G}} \right) \log_2 K}{(1 - \alpha)TK(\hat{\mu}_H - \hat{\mu}_L)}\nonumber 
\end{align}
where,
$\hat{\mu}_H = \hat{\mu}_{[1]} - (1-2/K)(\hat{\Delta}_{min} + \hat{\Delta}_{max})/2$, $\hat{\mu}_L = \hat{\mu}_{[1]} - (\hat{\Delta}_{min} + \hat{\Delta}_{max})/2$, $\hat\pi_{0,G} = \sigma_{out}(\hat\mu_G)/(\sigma_{in}(\hat{\mu}_G) + \sigma_{out}(\hat{\mu}_G))$, and $\hat\pi_{1,G} = \sigma_{in}(\hat\mu_G)/(\sigma_{in}(\hat{\mu}_G) + \sigma_{out}(\hat{\mu}_G))$. Here $\hat{\mu}_G$ is the sample average of the group pull. So finally, we aim to bound $\Delta_{\tau_G} := |\hat{\tau}_G - \tau_G|.$ Using triangle inequality, we write
\begin{align}
    |\hat{\tau}_G - \tau_G| \leq \underbrace{\left| \frac{\hat{\mu}_H + \hat{\mu}_L}{2} - \frac{\mu_H + \mu_L}{2} \right|}_{\text{(A): midpoint shift}} + \underbrace{\left| \frac{A_1}{B_1} - \frac{A_2}{B_2} \right|}_{\text{(B): LLR term shift}}\nonumber 
\end{align}
where, $A_1 = 2\sigma^2 \ln\left( \frac{\hat{\pi}_{0,G}}{\hat{\pi}_{1,G}} \right) \log_2 K$, $B_1 = (1 - \alpha)TK(\hat{\mu}_H - \hat{\mu}_L)$, $A_2 = 2\sigma^2 \ln\left( \frac{\pi_{0,G}}{\pi_{1,G}} \right) \log_2 K$, and $B_2 = (1 - \alpha)TK(\mu - \mu_L)$. Next, from triangle inequality, we have:
\begin{align}
    \left| \frac{\hat{\mu}_H + \hat{\mu}_L}{2} - \frac{\mu_H + \mu_L}{2} \right| &\leq \left| \frac{\hat{\mu}_H - \mu_H}{2}\right| + \left| \frac{\hat\mu_L - \mu_L}{2} \right| \nonumber
\end{align}
Note that $|\hat{\mu}_H - \mu_H| \leq |\hat{\mu}_{[1]} - \mu_{[1]}| + c_1|\hat{\Delta}_{min} - \Delta_{\min} | + c_2|\hat\Delta_{\max} - \Delta_{\max}|$, and  
$|\hat{\mu}_L - \mu_L| \leq |\hat{\mu}_{[1]} - \mu_{[1]}| + \Tilde{c}_1|\hat{\Delta}_{\min} - \Delta_{\min}| + \Tilde{c}_2|\hat{\Delta}_{\max} - \Delta_{\max}|$. Therefore,
\begin{align}
    \left| \frac{\hat{\mu}_H + \hat{\mu}_L}{2} - \frac{\mu_H + \mu_L}{2} \right| \leq C_1 \epsilon. \label{mean_deviation}
\end{align}
{Using high-probability bounds on $\hat{\mu}_{[1]}$, $\hat{\Delta}_{\min}$, and $\hat{\Delta}_{\max}$, and the constant $C_1$ depends on $K$.} Then, Using the triangle inequality again,
\begin{align}
&\left| \frac{A_1}{B_1} - \frac{A_2}{B_2} \right| 
\leq \frac{2\sigma^2\log_2(K)}{(1 - \alpha)TK} \nonumber \\
&\Bigg(\left| \frac{ \ln(\hat{\pi}_{0,G}/\hat{\pi}_{1,G}) - \ln(pi_{0,G}/\pi_{1,G}) }{ \mu_H - \mu_L } \right|
+ \nonumber \\
&|\ln(\hat{\pi}_{0,G}/\hat{\pi}_{1,G})| \left| \frac{1}{\hat{\mu}_H - \hat{\mu}_L} - \frac{1}{\mu_H - \mu_L} \right|\Bigg) \nonumber \\
&\leq \frac{2\sigma^2\log_2(K)}{(1 - \alpha)TK} \Bigg(\left| \frac{ \ln(\hat{\pi}_{0,G}/\pi_{0,G}) - \ln(\hat\pi_{1,G}/\pi_{1,G}) }{ \mu_H - \mu_L } \right| \nonumber \\
& +  \left|\frac{2\epsilon\ln(\hat{\pi}_{0,G}/\hat{\pi}_{1,G})}{(\mu_H-\mu_L)^2 - c_3\epsilon}\right| \Bigg) \nonumber \\
%&\leq  \frac{2\sigma^2\log_2(K)}{(1 - \alpha)TK}\Bigg(c_4\epsilon \left| \frac{ (2-\sigma_{out}(\hat{\mu}_G)-\sigma_{in}(\hat{\mu}_G)) }{ \mu_H - \mu_L } \right|
%+  \left|\frac{2\epsilon\ln(\hat{\pi}_{0,G}/\hat{\pi}_{1,G})}{(\mu_H-\mu_L)^2 - c_3\epsilon}\right| \Bigg) \nonumber \\
%&\leq \frac{2\sigma^2\log_2(K)}{(1 - \alpha)TK}\Bigg(c_4\epsilon \left| \frac{ (2-\sigma_{out}(\hat{\mu}_G)-\sigma_{in}(\hat{\mu}_G)) }{ \mu_H - \mu_L } \right|
%+  \left|\frac{2\epsilon(\ln(\pi_{0,G}/\pi_{1,G}) + c_5\epsilon(2-\sigma_{out}(\hat{\mu}_G)-\sigma_{in}(\hat{\mu}_G)))}{(\mu_H-\mu_L)^2 - c_3\epsilon}\right| \Bigg)  \nonumber \\
&\leq C_2\epsilon. \label{LLR_deviation} \nonumber
\end{align}
{Next, using similar high probability bound on $\hat{\mu}_{[1]}$, $\hat{\Delta}_{\min}$, and $\hat{\Delta}_{\max}$, and the constant $C_2$ depends on $K, T, \sigma^2, \alpha, \mu_{[1]}$ and  the sub-optimality gaps $\Delta_{\min}, \Delta_{\max}$.} Thus, combining \ref{mean_deviation},\ref{LLR_deviation}, we get $
    |\hat{\tau}_G - \tau_G| \leq C\epsilon$ for some $0 < C < \infty$. Accordingly, the probability of error in RE for each group is given by
\begin{align}
    &\mathcal{P}_{e,G} \leq \max\{\mathcal{P}_F, \mathcal{P}_M\} \leq  Q\bigg(\frac{\mu_H - \mu_L}{2} \sqrt{\frac{(1-\alpha)TK}{2\sigma^2\log_2(K)}}- \nonumber \\
    &\sqrt{\frac{2\sigma^2\log_2(K)}{(1-\alpha)TK}}\frac{\ln(\pi_{high,G}/\pi_{low,G})}{(\mu_H-\mu_L)} - \Tilde{C}\epsilon\bigg) 
\end{align}
and the total probability of error for all groups is bounded by
%\begin{align}
    $\delta_2 := \log_2(K) \mathcal{P}_{e,G}.$
%\end{align}
Therefore, finally we have $p^{RE}_G \leq \delta_1+ \delta_2$, which preserves the overall exponential decay of error with $T$ thanks to the exponential upper bound of the Gaussian Q-function and by selecting the weaker of the two terms constituting the summation. The case for the bounded distribution follows similarly.
}

\section{Experiments}
\label{app:experiments}
{ 
\subsection{Hardness Parameters and Upper Bounds}

\begin{figure*}[ht]
    \centering
    \subfloat[Arithmetic Gap.]{\includegraphics[width=0.25\textwidth]{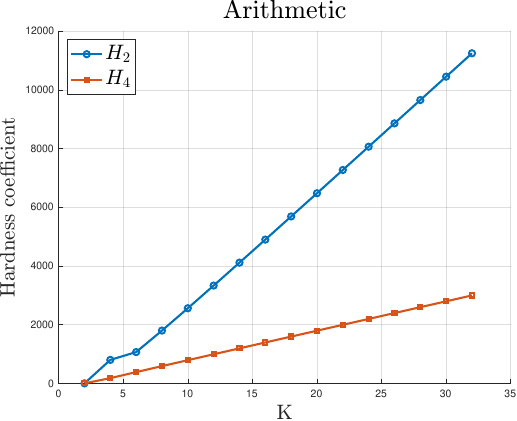}} 
    \subfloat[One real competitor.]{\includegraphics[width=0.25\textwidth]{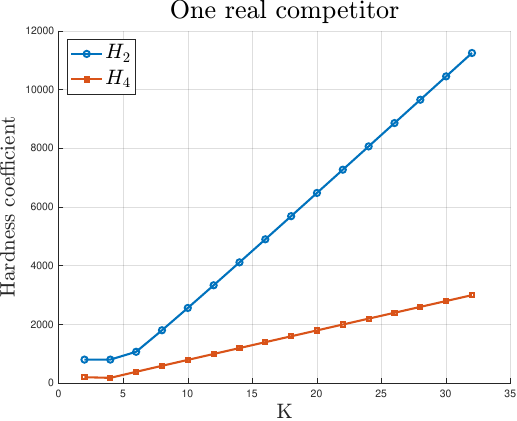}} 
    \subfloat[Two groups of competitors.]{\includegraphics[width=0.25\textwidth]{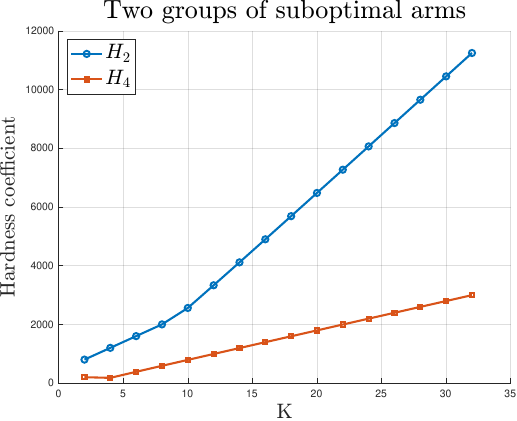}} 
    \subfloat[Single sub-optimality gap.]{\includegraphics[width=0.25\textwidth]{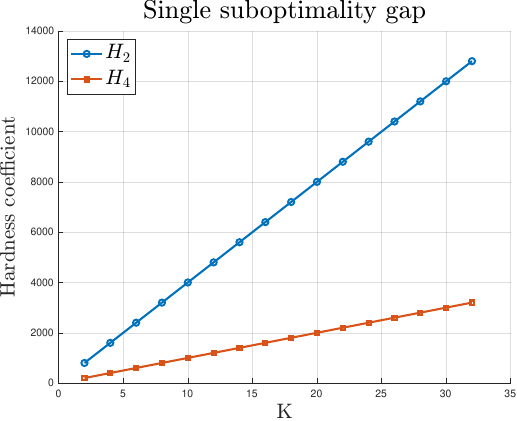}} \hfill
     \subfloat[$\Delta_{\min} = \Delta_{[1]}$, $K=4$.]{\includegraphics[width=0.25\textwidth]{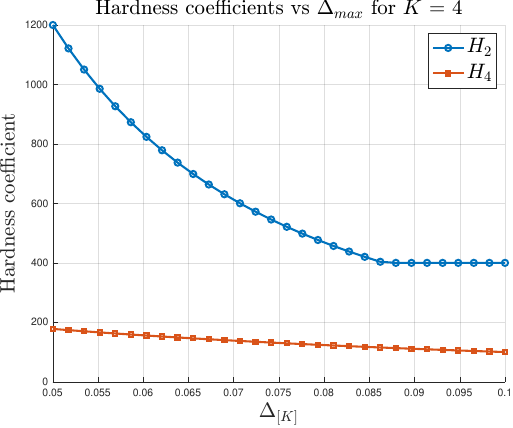}} 
    \subfloat[$\Delta_{\min} =\Delta_{[1]}$, $K=16$.]{\includegraphics[width=0.25\textwidth]{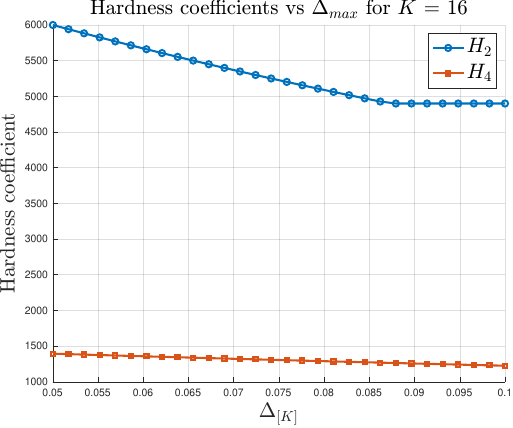}} 
    \subfloat[$\Delta_{\max} = \delta_{[K]}$, $K=4$.]{\includegraphics[width=0.25\textwidth]{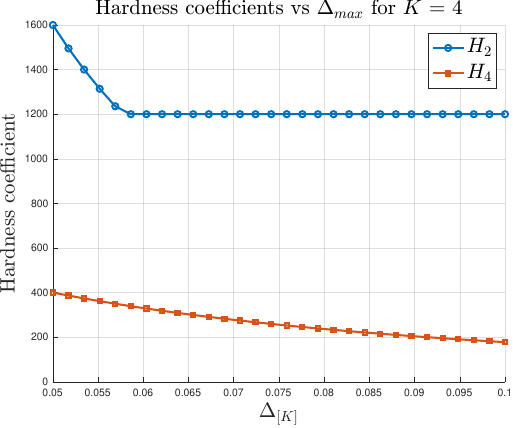}} 
    \subfloat[$\Delta_{\max} = \Delta_{[K]}$, $K=16$.]{\includegraphics[width=0.25\textwidth]{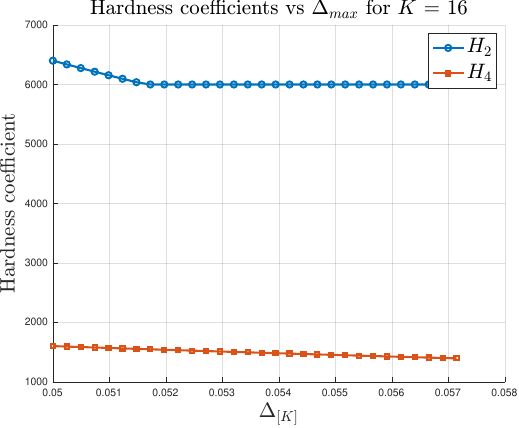}}
    \caption{Variation of the hardness parameters $H_2$ and $H_4$ with the number of arms and the sub-optimality gaps. Arithmetic gap refers to equally spaced sub-optimality gaps between $\Delta_{\min}$ and $\Delta_{\max}$. One real competitor refers to the case where $\Delta_{[1]}$ is small while $\Delta_{[i]}, i\neq 1$ are large. Two groups of competitors refer to an equal number of arms having a sub-optimality gap near $\Delta_{[min]}$ and $\Delta_{\max}$ respectively. Finally, single sub-optimality gap refers to the case where $\Delta_{[i]}$ is the same for all $i$.}
    \label{fig:h2h4}
\end{figure*}
\begin{figure*}[ht]
    \centering
    \subfloat[ Arithmetic gap.]{\includegraphics[width=0.25\textwidth]{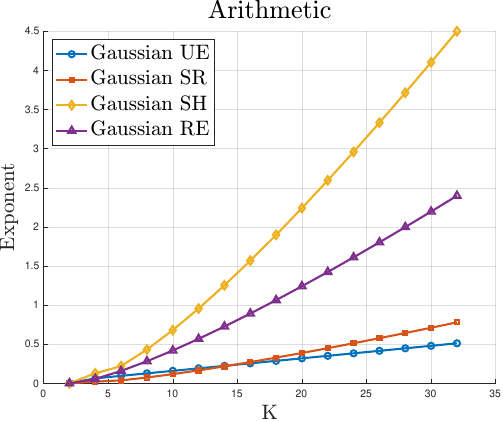}} 
    \subfloat[One real competitor.]{\includegraphics[width=0.25\textwidth]{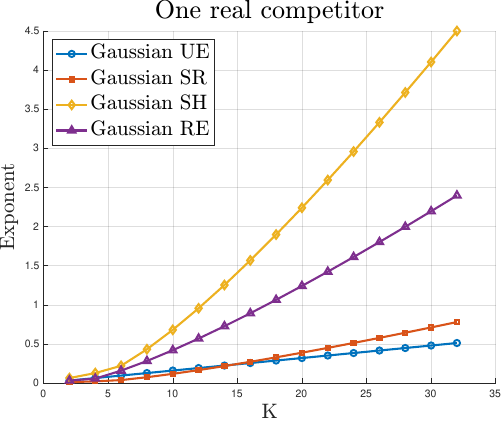}} 
    \subfloat[Two groups of competitors ]{\includegraphics[width=0.25\textwidth]{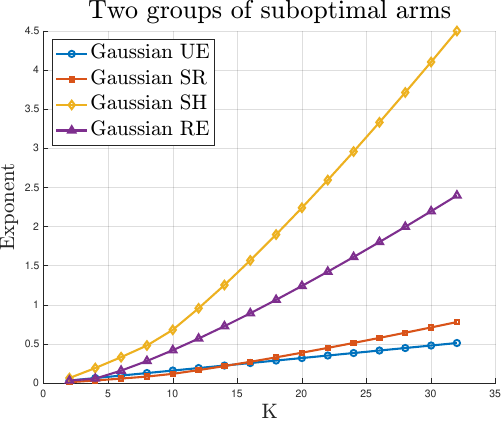}} 
    \subfloat[Single sub-optimality gap.]{\includegraphics[width=0.25\textwidth]{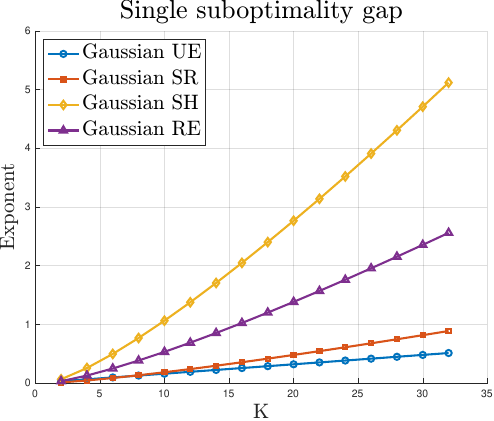}} \hfill
     \subfloat[$\Delta_{\min} = \Delta_{[1]}$, $K=4$.]{\includegraphics[width=0.25\textwidth]{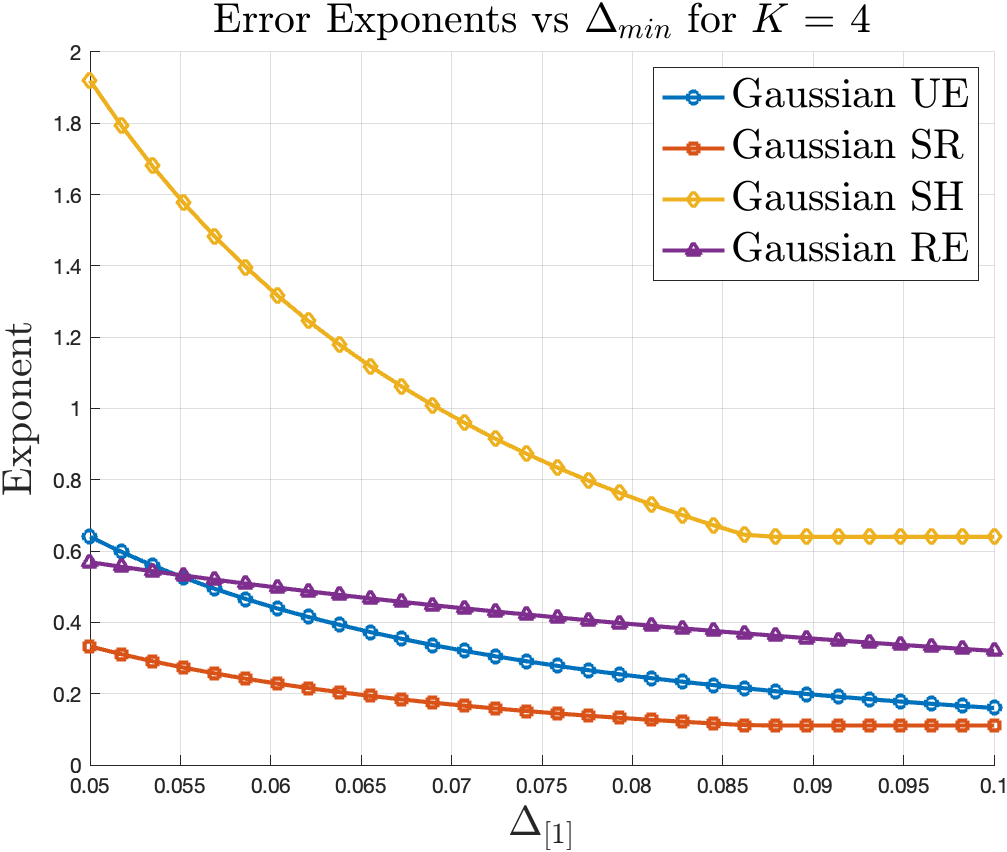}} 
    \subfloat[$\Delta_{\min} = \Delta_{[1]}$, $K=16$.]{\includegraphics[width=0.25\textwidth]{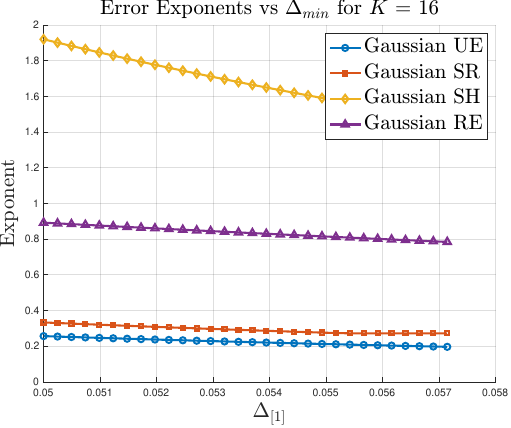}} 
    \subfloat[$\Delta_{\max} = \Delta_{[K]}$, $K=4$.]{\includegraphics[width=0.25\textwidth]{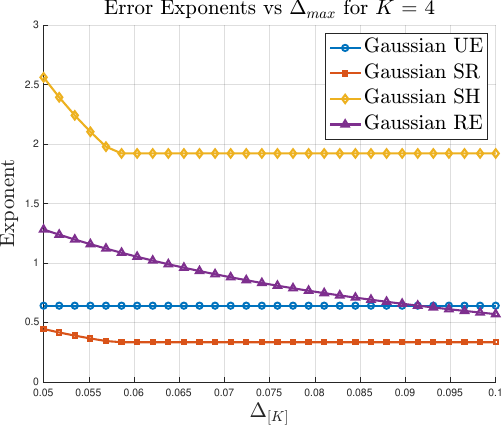}} 
    \subfloat[$\Delta_{\max}  = \Delta_{[K]}$, $K=16$.]{\includegraphics[width=0.25\textwidth]{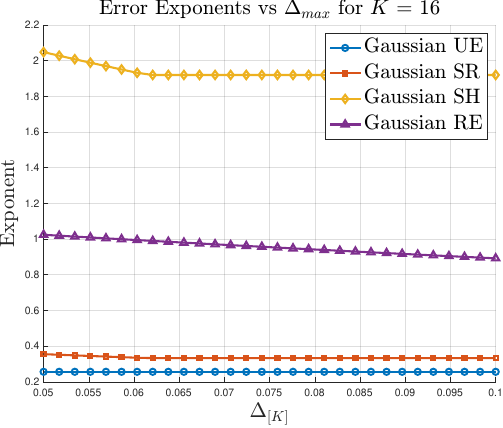}}
    \caption{The exponential term of the error bounds for Gaussian rewards. Note that is the argument of the negative exponent, hence, the higher the better.}
    \label{fig:GR}
\end{figure*}
In this section, we quantitatively discuss the distinction between the hardness parameters $H_2$ and $H_4$. Note that from the separability conditions, i.e, \(
\Delta_{[1]} \leq \Delta_{[K]} \leq \frac{\Delta_{[1]}}{1 - 2/K}\), the following bounds hold true.
\begin{align}
&\frac{1}{2\Delta_{[1]}} \geq \frac{1}{\Delta_{[1]} + \Delta_{[K]}} \geq \frac{1}{2\Delta_{[K]}} \nonumber \\
\Rightarrow &\frac{1}{(\Delta_{[1]} + \Delta_{[K]})^2} \in \left[\frac{1}{4\Delta_{[K]}^2}, \frac{1}{4\Delta_{[1]}^2} \right].
\end{align}
Furthermore, recall that $\tilde{H}_4 := K H_4$, and consequently,
\begin{align}
\frac{(K-2)^2}{K \Delta_{[1]}^2} \leq 4 \tilde{H}_4 \leq \frac{K}{\Delta_{[1]}^2}.
\end{align}

Now to place the comparison of $H_2$ and $\tilde{H}_4$ on the same page, we note that
\begin{align}
\frac{H_1}{\log(2K)} \leq H_2 \leq \frac{K}{\Delta_{[1]}^2}, \quad \text{and} \quad H_1 = \sum_i \frac{1}{\Delta_{[i]}^2} \geq \frac{K}{\Delta_{[K]}^2}. \nonumber
\end{align}
Thus, under separability, we observe that:
\begin{align}
{\frac{(K-2)^2}{K\log_2(2K)\Delta_{[1]}^2} \leq H_2 \leq \frac{K}{\Delta_{[1]}^2}},\; \text{and} \quad {4\tilde{H}_4 \approx \Theta\left(\frac{K}{\Delta_{[1]}^2}\right)}. \nonumber
\end{align}

For most real-world systems this lower bound on $H_2$ is achieved corresponding to the lower bound on $H_1$, i.e., when just one arm has a sub-optimality gap $\Delta_{[1]}$ and all other arms have a gap $\Delta_{[K]}$. In this case, $H_2$ will either be $K/\Delta_{[1]}^2$, or $K/\Delta_{[K]}^2$, or $j/\Delta_{[1]}^2$ if the second best arm has an index $j > K-4+4/K$. That is, both $H_2$ and $\tilde{H}_4$ scale as $\Theta(K / \Delta_{[1]}^2)$, with different constant factors reflecting the structure of the gap profile. Let us discuss the variation $H_2$ and $H_4$ for various cases in Fig.~\ref{fig:h2h4}. These plots reveal that $H_4$ is often smaller than $H_2$, especially in regimes with many well separated arms. Since $H_4$ governs the sample complexity of RE and appears in its exponent, this supports our claim that RE can achieve lower error probabilities in such structured settings. Recall that in the upper bound of the algorithms consist of two terms one leading term and one exponential. The leading term has a clear advantage in the case of RE and SH since they are logarithmic in $K$ while the SR is quadratic in $K$. The RE has a further advantage of a constant $3$ with respect to SH. We further show below that how RE and SH both have superior advantage in the exponential term. Albeit the SH outperforms the RE in terms of the exponential term in general, the choice of the algorithm may depend on the exact value of $K$. The comparison of the exponential terms of different algorithms are presented in Fig.~\ref{fig:GR} for Gaussian rewards. The observations for Bernoulli rewards were similar and hence we skip it for this discussion. As we shall see in the next sections, in many structured instances (e.g., well-separated groups and low variance), RE also achieves a sharper slope. We combine these observations in the following remark.
\begin{remark}
    As compared to SR (quadratic in $K$), the leading terms of SH and RE (logarithmic in $K$) have a clear advantage. In addition, the leading term of RE has a factor 3 advantage over that of SH. The  exponent term of the SH appears to be the strongest followed by RE. Due to the trade-off between the leading terms and the exponential term, an appropriate choice of algorithm depends on the exact value of $K$ and the sub-optimality gaps.
\end{remark}
}

\begin{figure}[ht]
    \centering
    \subfloat[\centering$K=8$, One Real Competitor, Gaussian Rewards \\
$\Delta = 0.05$, $\sigma^2 = 0.5$, $H_1 = 2800$]
{\includegraphics[width=0.22\textwidth]{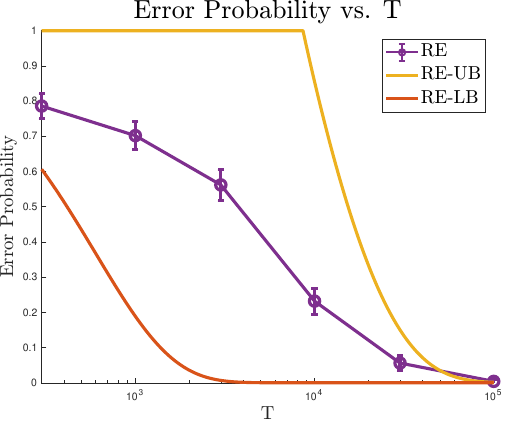}} 
\hfil
    \subfloat[\centering$K=1024$, Single sub-optimality Gap, Gaussian Rewards
$\Delta = 0.5$, $\sigma^2 = 0.1$, $H_1 = 4092$]{\includegraphics[width=0.22\textwidth]{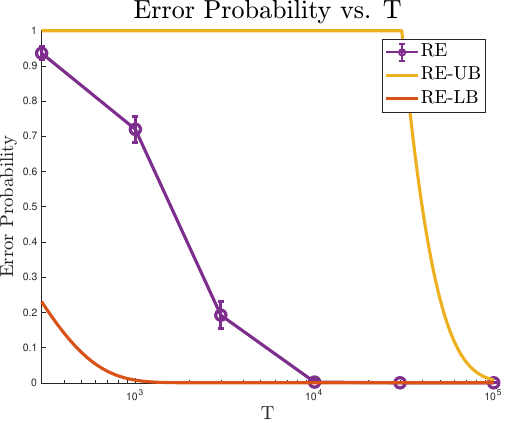}}
\hfil
\subfloat[\centering$K=8$, One Real Competitor, Bernoulli Rewards\\
$\mu^* = 0.5$, $\Delta_{[1]} = 0.1$, $H_1 = 1420$]{\includegraphics[width=0.22\textwidth]{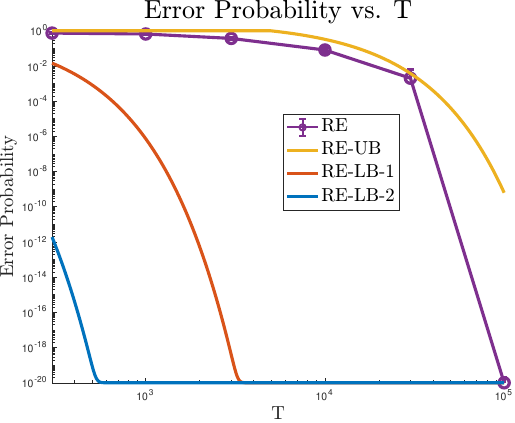}} 
\hfill
%     \subfloat[\centering$K=512$, Single sub-optimality Gap, Bernoulli Rewards
% $\mu^* = 0.9$, $\Delta = 0.8$, $H_1 = 792$]{\includegraphics[width=0.45\textwidth]{plots/bernoulli_lb2.pdf}} 
    \subfloat[\centering$K=512$, Single sub-optimality Gap, Bernoulli Rewards
$\mu^* = 0.9$, $\Delta = 0.8$, $H_1 = 792$]{\includegraphics[width=0.22\textwidth]{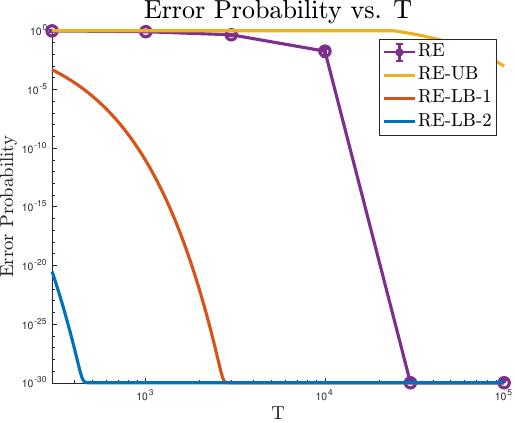}} 
\caption{Comparison with bounds.}
\label{fig:Gauss}
\end{figure}

% \begin{figure}[ht]
%     \centering
% %     \subfloat[\centering$K=8$, One Real Competitor, Bernoulli Rewards\\
% % $\mu^* = 0.5$, $\Delta_{[1]} = 0.1$, $H_1 = 1420$]{\includegraphics[width=0.45\textwidth]{plots/bernoulli_lb1.pdf}} 
% \subfloat[\centering$K=8$, One Real Competitor, Bernoulli Rewards\\
% $\mu^* = 0.5$, $\Delta_{[1]} = 0.1$, $H_1 = 1420$]{\includegraphics[width=0.45\textwidth]{plots/bernoulli_lb1_log.pdf}} 
% \hfill
% %     \subfloat[\centering$K=512$, Single sub-optimality Gap, Bernoulli Rewards
% % $\mu^* = 0.9$, $\Delta = 0.8$, $H_1 = 792$]{\includegraphics[width=0.45\textwidth]{plots/bernoulli_lb2.pdf}} 
%     \subfloat[\centering$K=512$, Single sub-optimality Gap, Bernoulli Rewards
% $\mu^* = 0.9$, $\Delta = 0.8$, $H_1 = 792$]{\includegraphics[width=0.45\textwidth]{plots/bernoulli_lb2_log.pdf}} 
% \caption{Comparison with the bounds - Bernoulli rewards.}
% \label{fig:Bern}
% \end{figure}

\subsection{Numerical Simulations}
\begin{figure}[ht]
    \centering
    \subfloat[$K=4$, Arithmetic Gap.]{\includegraphics[width=0.45\linewidth]{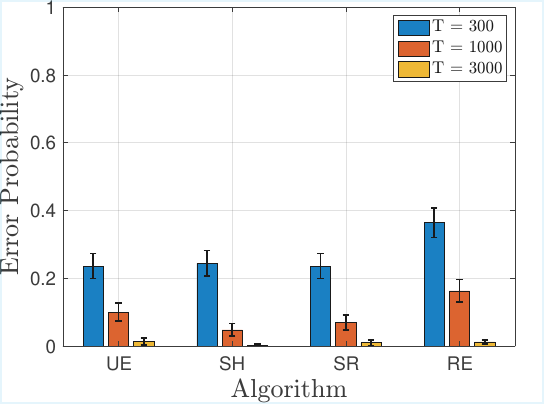}} 
    \subfloat[$K=8$, Arithmetic Gap.]{\includegraphics[width=0.45\linewidth]{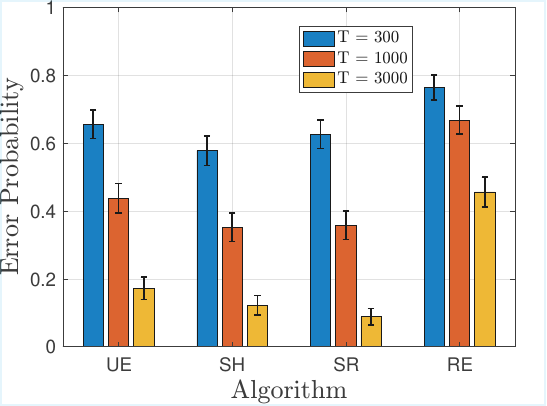}} 
    \hfill
    \subfloat[$K=8$, One real competitor.]{\includegraphics[width=0.45\linewidth]{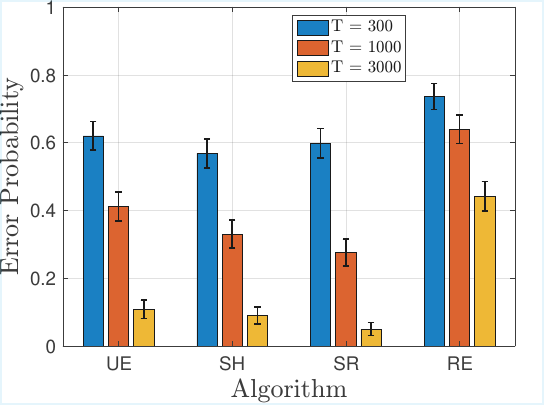}} 
    \subfloat[$K=8$, 2 groups of competitors.]{\includegraphics[width=0.45\linewidth]{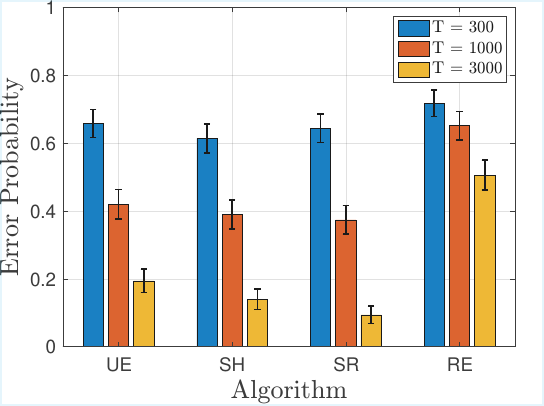}}
    \caption{\centering Probability of error for $\Delta_{[1]} = 0.05$ and $\sigma^2 = 0.5$ for Gaussian rewards.}
     \label{fig:res1}
\end{figure}

\begin{figure}[ht]
    \centering
    \subfloat[$K=4$, Arithmetic Gap.]{\includegraphics[width=0.45\linewidth]{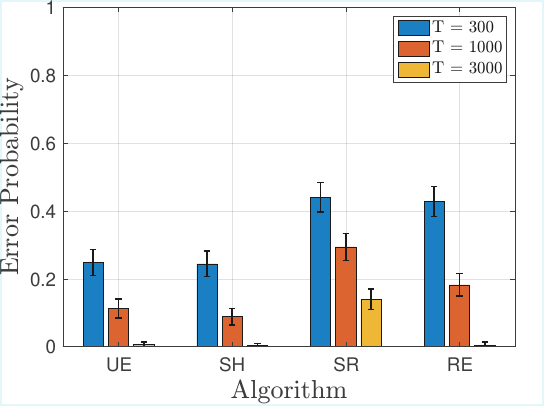}} 
    \subfloat[$K=8$, Arithmetic Gap.]{\includegraphics[width=0.45\linewidth]{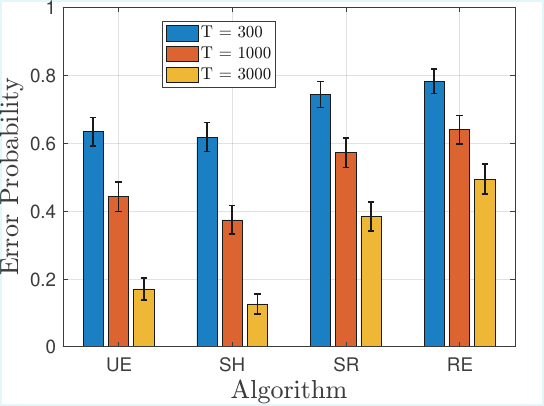}} 
    \hfill
    \subfloat[$K=8$, One real competitor.]{\includegraphics[width=0.45\linewidth]{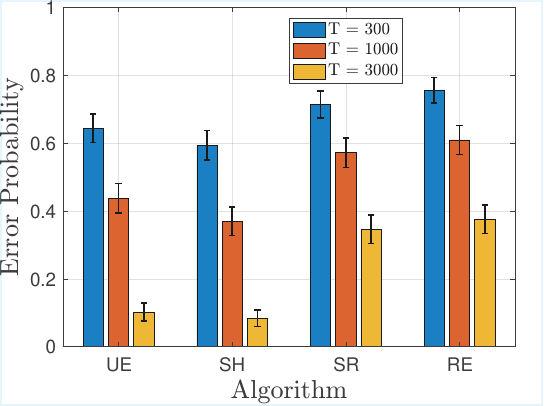}} 
    \subfloat[$K=8$, 2 groups of competitors.]{\includegraphics[width=0.45\linewidth]{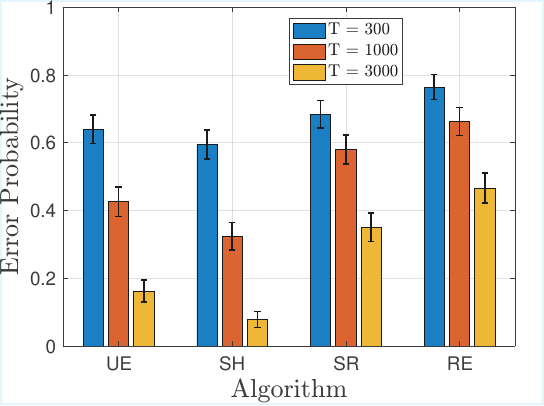}} 
    \caption{\centering Probability of error for $\Delta_{[1]} = 0.05$ and $\mu^* = 0.5$ for Bernoulli rewards.}
          \label{fig:res2}
\end{figure}

\begin{figure}[ht]
    \centering
    \subfloat[\centering$K=1024$, Single sub-optimality Gap, Gaussian Rewards \\
    $\Delta = 0.5, \sigma^2 = 0.1$]{\includegraphics[width=0.45\linewidth]{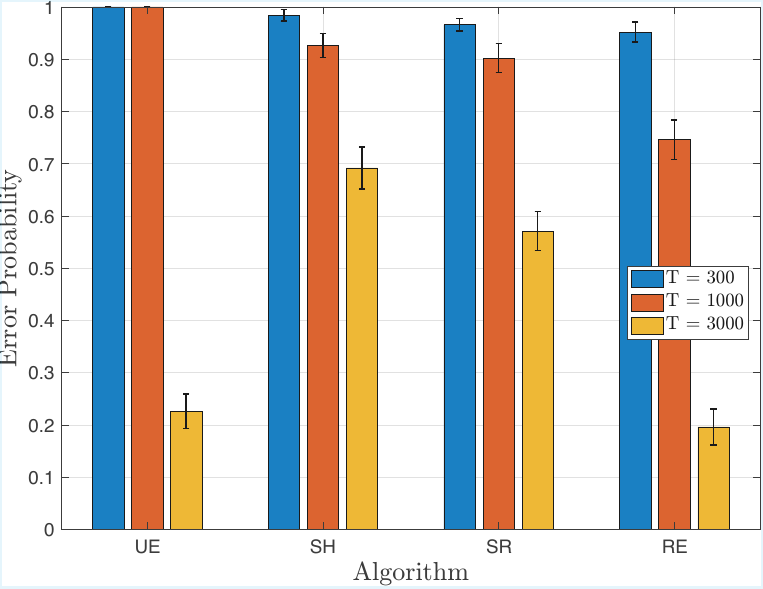}} 
    \subfloat[\centering$K=512$, Single sub-optimality Gap, Bernoulli Rewards\\
    $\mu^* = 0.9, \Delta=0.8$]{\includegraphics[width=0.45\linewidth]{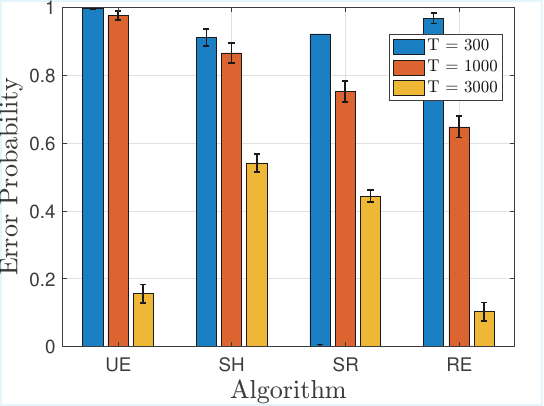}} 
    \hfill
    \caption{\centering Probability of error for large $K$, small variance case}
    \label{fig:res3}
\end{figure}

{ We observe from Fig.~\ref{fig:Gauss} that the empirical error probability of the RE algorithm does not exactly match the theoretical lower bounds, but tracks them reasonably well under certain conditions. In particular, in the Gaussian setting with large sub-optimality gap and low variance, the performance of RE closely approximates the lower bound for budgets as low as approximately $2H_1$. Notably, this occurs despite the problem involving a much larger number of arms, demonstrating that RE effectively scales with $K$ in this regime.
In contrast, when the sub-optimality gap is small and variance is high, RE's error decays more slowly. This is expected, as identifying the best arm under fine-grained reward differences is inherently harder, and aligns with our results.

In the case of bounded rewards, the theoretical lower bounds decay extremely rapidly with $T$, resulting in nearly vanishing values even for moderate budgets. This can be seen clearly in the log-scale plots. As a result, the empirical performance of RE appears significantly above the lower bounds in absolute terms. However, we observe that RE still exhibits consistent exponential decay. This suggests that while the bounds may not be tight in practically realizable regimes of $T$, RE maintains asymptotically optimal behavior in terms of convergence rate.}

{ We examine the variation of the probability of error with different system parameters. The trends with respect to the number of arms ($K$) evaluates how RE scales with problem size. Then, the variation with reward distribution analyzes the robustness of RE across different statistical model. The sample efficiency is outlined with respect to varying budget. In each case, we conduct 500 independent trials and report the empirical mean along with $95\%$ confidence intervals, thereby quantifying the variability of performance.

The synthetic results of Figs.~\ref{fig:res1} - \ref{fig:res3} show that RE under-performs compared to its counterparts (SR, SH, and UE) in some problem instances. This is attributed to the inherent hardness of the instance and the high variance in the group-level observations. In such cases, the aggregate feedback becomes ambiguous, thereby making the decoding step more error-prone. This highlights the necessity of the separability assumption without sufficient separation between the optimal and sub-optimal arms, the theoretical advantages of RE do not manifest empirically. To this end, we experiment with a well separated problem instance with large $K$ and low observation noise. In this regime, the fixed schedules of SR and SH fare poorly as the initial phases will be allotted 0 budget and the algorithm will have to eliminate random arms. The UE algorithm also performs poorly until the budget is at least $K$, whereas RE does not have any such drawbacks due to it's logarithmic nature. To further elaborate on this claim, let us discuss two case studies in the next section.}

\section{Case Studies}

{ 
\subsection{Jammer Design}
\begin{figure}
    \centering
    \includegraphics[width=0.75\linewidth]{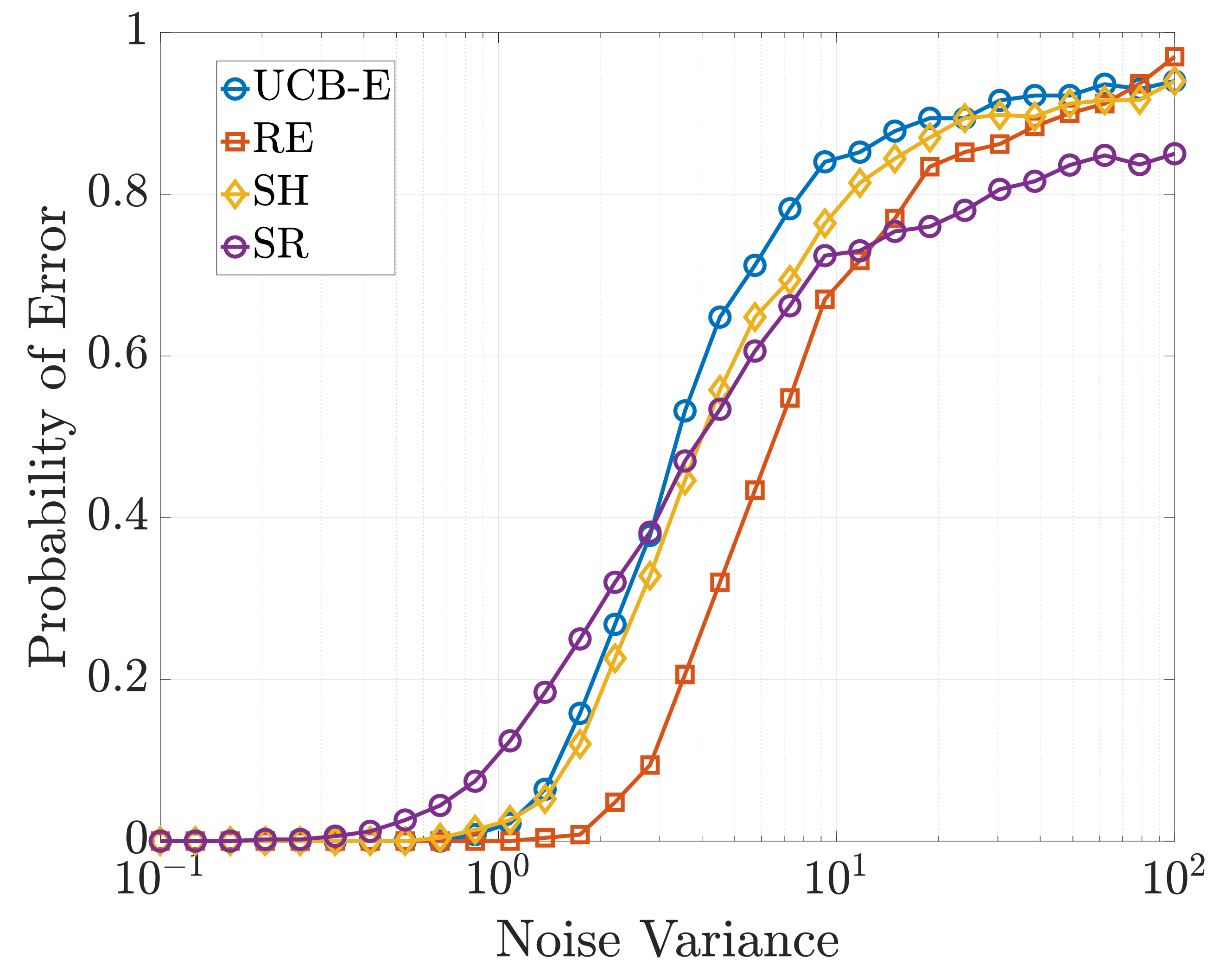}
    \caption{Probability of error vs noise variance.}
    \label{fig:jamming}
\end{figure}
In modern electronic warfare and cognitive radio environments, a jammer aims to disrupt a communication or radar receiver by transmitting signals that maximally interfere with the receiver's detection process. In many systems, the receiver employs a matched filter designed to detect a known or fixed waveform from a finite set of orthogonal templates. These templates are often derived from orthogonal codes, frequency tones, or time-shifted pulse sequences. We study the efficacy of the proposed BAI framework in selecting the best jammer sequence from the waveform codebook.

In this regard, consider a jammer equipped with a finite set of orthogonal baseband waveforms $\mathcal{W} := \{w_i(t)\}, t \in [K]$, where each $w_k(t)$ is a real-valued or complex-valued waveform of unit energy. i.e.,  $\int |w_k(t)|^2 \, dt = 1, \quad \forall k = 1, \dots, K$, and $\langle w_j, w_k \rangle = 0$ for all $j \ne k$. The target receiver is assumed to use a matched filter $h(t) = w^*(t)$ tuned to an unknown waveform $w^* \in \mathcal{W}$ for detection. This waveform represents the expected signal-of-interest, such as a radar pulse or control pilot signal. %In fact, based on the exact application, the library of the candidate waveforms may differ. For example,  in wireless communications (e.g., LTE/WiFi jamming) the waveforms may be OFDM sequences~\cite{litwin2001principles}, Zadoff-Chu sequences~\cite{andrews2022primer}, or Walsh codes~\cite{ho2014high}. On the contrary, radar applications time shifted chirps or Hermite pulses~\cite{de2003design}. Military applications may include Walsh, Gold/Kasami codes~\cite{shi2019new}.
At each time instant (or round) $t$, the jammer selects a subset $S_t \subseteq \{1, 2, \dots, K\}$ of $|S_t| = K/2$ waveforms as per the RE algorithm to transmit, and sends their normalized linear superposition, i.e., $x_t(t) = \sqrt{\frac{2}{K}} \sum_{k \in S_t} w_k(t)$. This normalization ensures constant total transmit power, maintaining practical power constraints. The received signal at the target receiver, in the absence of its own transmissions, is modeled as
%\begin{align}
    $y_t(t) = x_t(t) + n_t(t)$,
%\end{align}
where $n_t(t)$ is additive white Gaussian noise (AWGN) with power spectral density $N_0/2$ per dimension. Then, the matched filter output at the receiver for waveform $w_j$ is
%\begin{align}
    $r_j(t) = \int y_t(\tau) w_j^*(\tau) \, d\tau.$
%\end{align}
Thanks to orthogonality, we have
%\begin{align}
    $r_j(t) = \sqrt{\frac{2}{K}} \sum_{k \in S_t} \langle w_k, w_j \rangle + \eta_j(t)$,
%\end{align}
where $\eta_j(t) \sim \mathcal{N}(0, N_0)$ due to noise projection onto $w_j$. Assuming the receiver only responds to $w^* = w_{j^*}$, and the jammer can measure the output power at that matched filter (e.g., via signal leakage or energy sensing), the jammer observes a scalar reward
%\begin{align}
   $R_t = |r_{j^*}(t)|^2$. 
%\end{align}
Again, due to orthogonality, this simplifies to
\begin{align}
    R_t = \begin{cases}
\frac{2}{K} + \text{noise}, & \text{if } j^* \in S_t, \\
0 + \text{noise}, & \text{otherwise}.
\end{cases}
\end{align}

The jammer does not know the identity of $j^*$ a priori and must infer it through interactions. At each round, the jammer selects a subset $S_t$ and receives a noisy scalar reward
%\begin{align}
    $R_t = \frac{1}{m} \mathbb{I}\{j^* \in S_t\} + \xi_t$,
%\end{align}
where $\xi_t \sim \mathcal{N}(0, \sigma^2)$ models additive observation noise. The learning problem is thus to identify the waveform index $j^*$ with the highest mean reward using the fewest number of averaged subset queries. This forms a special case of a combinatorial bandit problem with large sub-optimality gap and aggregate feedback. Now, to put this into our framework, let $\mu_k = \mathbb{E}[R_t \mid S_t = \{k\}]$. Then, $\mu_k = 1$ only for the optimal arm and 0 otherwise.
% \begin{align}
%     \mu_k = \begin{cases}
% 1 + \mathbb{E}[\xi_t], & \text{if } k = j^*, \\
% 0 + \mathbb{E}[\xi_t], & \text{otherwise}.
% \end{cases}
% \end{align}
Hence, the sub-optimality gap $\Delta = \mu_{j^*} - \max_{k \ne j^*} \mu_k = 1$, which makes the problem suitable to apply our framework.

We test for a case with $K = 16$ waveforms and with varying noise power. In Fig.~\ref{fig:jamming} we show for limited noise variance, the jammer is able to identify the best arm more efficiently. However, as the noise variance increases, grouping more arms together for combinatorial exploration is detrimental. As a result the the probability of error increases more rapidly for RE as compared to that of the other algorithms and eventually the performance of RE drops to become the worst. This highlights the necessity of the assumption of separability as well as the limitations of the grouping strategy.

\subsection{Detection of Active Radar Channel}
We consider a wideband active radar channel detection problem where a radar transmits in one of $K$ known channels. The objective of the detector is to detect the active channel using an energy detection mechanism. Each channel is observed by collecting complex I/Q samples, and the problem is cast as a \ac{BAI} problem, where each channel corresponds to an arm. For emulating the single radar transmission data, we leverage the RadChar-Tiny dataset~\cite{radchar} in a hypothetical scenario with 8 channels. While the radar signal is present in one of the 8 channels, the other channels only contain noise. We simulate the noise of the other channels by computing the baseline noise floor of the dataset (obtained by considering the periods where the radar pulse is absent). Indeed, for a more realistic assessment for wideband radar detection, sophisticated datasets such as the recently introduced RadDet~\cite{huang2025raddet} can be employed. Since our focus is to demonstrate the efficacy of the \ac{BAI} to detect a single active channel, we select the simpler preprocessed variant of RadChar dataset.

Let $f_s = 3.2$ MHz denote the sampling rate and $T$ be the dwell time for each observation. Each play of an arm involves collecting $N = T \cdot f_s$ samples from that channel. Let $x_k[n] \in \mathbb{C}$ denote the complex baseband I/Q sample from channel $k$ at sample index $n \in \{0, 1, \dots, N-1\}$. The radar signal parameters consists of a uniformly sampled number of pulses in the range 2 to 6, uniformly sample pulse width in $[10,16]$ $\mu$s, uniformly sampled pulse repetition interval in the range 17 to 23 $\mu$s, and a uniform pulse delay between 1 to 10 $\mu$s. For further details on the characteristics of the radar signal in this data set, please refer to \cite{radchar}.
    
Now, from the BAI perspective, {\it playing an arm} corresponds to sampling a channel for a duration $T$ followed by energy detection. The energy computed in channel $k$ is:
\begin{align}
E_k = \sum_{n=0}^{N-1} |x_k[n]|^2 = \sum_{n=0}^{N-1} |x_I[n]|^2 + |x_Q[n]|^2,
\end{align}
where $x_I[n]$ and $x_Q[n]$, respectively are the I and Q samples of the dataset. Then, $E_k$ is the reward that the player gets in sampling the channel $k$. With a total time (i.e., samples) $T_{\rm tot}$, the number of plays allowed to the player thus is $\frac{T_{\rm tot}}{T}$. Specifically, the observed energy $E_k^{(t)}$ at round $t$ and $\mu_k = \mathbb{E}[E_k]$. As per our formulation, for $k \neq k^\star$, the mean is $\mu_k = N \sigma^2$ where $\sigma^2$ is the variance of the noise in the dataset. On the other hand, for $k = k^\star$, $\mu_k > N \sigma^2$ due to the presence of the radar signal. The goal is to identify $k^\star = \arg\max_k \mu_k$, i.e., the channel with the highest expected energy.

\begin{table*}[ht]
\centering
\begin{tabular}{|c|c|c|c|c|c|}
\hline
Time of observation& Bandit Plays & {SH}& {SR}& {RE (without $\mu_{[1]}$)}&{RE (with $\mu_{[1]}$)}\\
\hline
35ms&1200 & 0.07& 0.23& 0.28 &0.001\\
90ms&3000 & 0.006& 0.06& 0.07 &0\\
180ms&6000 & 1e-4& 0.01& 0.008 &0\\
\hline
\end{tabular}
\caption{Probability of error for $K=8$ with different algorithms.}
\label{radartable}
\end{table*}
According to our multi-play \ac{BAI} framework, playing multiple arms corresponds to sampling multiple bands simultaneously and observing the power of the combined signal power in all these channels. The trade-off of the multi-play \ac{BAI} naturally carries forward to this application - sensing multiple channels together increases the variance of the signals but provides a wider view of the scenario. We employ the proposed RE algorithm to this problem and report the results in Table~\ref{radartable}. Intuitively, longer the observation window, the lower is the probability of error in detecting the active channel. Interestingly, we observe significant improvement in the case where the player has prior knowledge of $\mu_{[1]}$, i.e., when the priors can be exactly calculated and provided to the player before the start of the play and the case where the priors need to be engineered following an estimation of $\mu_{[1]}$ during the initial exploration phase. From the perspective of radar detection the knowledge of $\mu_{[1]}$ corresponds to the knowledge of the radar signature and network geometry (e.g., relative distance of the transmitter to the receiver, propagation environment, etc.). This observation leads us to conclude the following remark.
\begin{remark}
    In case the sub-optimality gaps are known the player and the problem is merely that of identifying the best arm, empirically RE outperforms other algorithms, while its performance is limited when this sub-optimality gaps are to be estimated as a sub-routine of the algorithm.
\end{remark}

}

\section{Conclusions and Future Work}
We proposed an arm grouping strategy for the fixed budget BAI problem where the agent is allowed to sample multiple arms at a time. The agent has to trade-off between sampling a larger number of arms, thereby obtaining a wider view of the environment at the cost of reduced information about per-arm reward distributions, or playing fewer arms that reveal fine-grained reward distributions at the cost of increasing the sample complexity. Our algorithm is based on pulling groups of arms formed using binary representation of the arm indices and Hamming codes. For each group the agent performs a \ac{LRT} to determine whether or not the best arm is present in that group. For Gaussian and bounded rewards, we derived the upper bound of the probability of error and discussed the conditions under which it outperforms the state-of-the-art algorithms for the single pull setting. The case studies highlight that the choice of an algorithm for the combinatorial case is not straightforward since the advantage of combinatorial pulls is limited in case the rewards are not separable.

To the best of our knowledge this is the first work that addresses the BAI problem with combinatorial exploration where the rewards are non-trivial (e.g., not constant). We considered a special case of this framework where on sampling multiple arms, the agent receives the sample average of the individual rewards of the arms pulled. Other functions than the sample average of the rewards can be of interest based on the application. Furthermore, deriving the lower bound of the probability of error with combinatorial pulls is indeed an interesting open question that we will address in a future work.

\appendices
\section{Analysis of the Uniform Exploration strategy}
\label{app:UE}
In this section, we give the results for the uniform exploration strategy for the best-arm identification problem in the non-combinatorial setting. Let the means of the $K$ arms be $(\mu_1, \mu_2, \dots, \mu_K)$ and the budget be $T$. We allot equal budget to all the arms and pull each arm $T/K$ times. The algorithm makes a mistake if any of the suboptimal arms has a higher empirical mean after $T/K$ rounds. Therefore,

\begin{align}
    p_{(\cdot)}^{\rm UE} &= \bP_{\nu}(\hat{a}_T \neq a^*) = \bP_{\nu}(\cup_{i = 1, i \neq a^*}^K \hat{X}_{a^*,T/K} < \hat{X}_{i,T/K}) \nonumber\\
        &\leq \sum_{i=1, i\neq a^*}^K \bP_{\nu}(\hat{X}_{a^*,T/K} < \hat{X}_{i,T/K}) \nonumber\\
        &\leq \sum_{i=2}^K \bP_{\nu}((\hat{X}_{i,T/K} - \mu_{[i]}) - (\hat{X}_{a^*, T/K} - \mu_{[1]}) > \Delta_{[i]}). \nonumber
\end{align}

Thus, for Gaussian bandit models with variance of all arms being $\sigma^2$, we have
\begin{equation}
    (\hat{X}_{i,T/K} - \hat{X}_{a^*,T/K}) \sim \cN(\mu_{[i]}-\mu_{[1]}, \frac{2K\sigma^2}{T}) \nonumber
\end{equation}
\begin{align}\label{eqn:uniform_exploration_bound_Gaussian}
    p_{G}^{\rm UE} &\leq \sum_{i=2}^K \bP_{\nu}\bigg(\cN(0,1) > \Delta_{[i]}\sqrt{\frac{T}{2K\sigma^2}}\bigg) \nonumber\\
    &\leq \sum_{i=2}^K Q\bigg(\Delta_{[i]}\sqrt{\frac{T}{2K\sigma^2}}\bigg) \leq (K-1)Q\bigg(\Delta_{[1]}\sqrt{\frac{T}{2K\sigma^2}}\bigg) \nonumber\\
    &\leq (K-1)\sqrt{\frac{H_3\sigma^2}{\pi T}} \exp\bigg(\frac{-T}{4H_3\sigma^2}\bigg).\nonumber
\end{align}

For bounded bandit models, instead of using $Q$ functions, we make use of Hoeffding's inequality to bound the error probability as,
\begin{align}
    p_{B}^{\rm UE} &\leq \sum_{i=2}^K \bP_{\nu}((\hat{X}_{i,T/K} - \mu_{[i]}) - (\hat{X}_{J^*, T/K} - \mu_{[1]}) > \Delta_{[i]}) \nonumber\\
    &\leq \sum_{i=2}^K \bP_{\nu}\bigg( \sum_{t=1}^{T/K} X_{i,t} - X_{J^*,t}  - \frac{T}{K}\bigg(\mu_{[i]} - \mu_{[1]}\bigg)> \frac{T}{K}\Delta_{[i]}\bigg) \nonumber\\
    &\leq \sum_{i=2}^K \exp\bigg(\frac{-T\Delta_{[i]}^2}{2K}\bigg)  \leq (K-1)\exp\bigg(\frac{-T}{2H_3}\bigg).  \nonumber
\end{align}

\bibliographystyle{IEEEtran}
\bibliography{refer.bib}

% \section{On the distribution of $\mu_H$ and $\mu_L$}\label{app:mu_H,mu_L distribution}

% We assume that the best arm is chosen to be one of the $K$ arms at random. 
% We now plot the distribution of the conditional means $\mu_H, \mu_L$ of the group with and without the best arm.
% \begin{figure}
%     \centering
%     \includegraphics[width=0.7\linewidth]{figs/distribution.pdf}
%     \caption{The conditional distribution of $\mu_H$ and $\mu_L$}
%     \label{fig:enter-label}
% \end{figure}
% The above figure is for $K = 8$ and $\Delta_{[K]} = \frac{K}{K-2}\Delta_{[2]}$. 
% For $\Delta_{[2]} = \Delta_{[K]}$, the conditional means are point masses at $\mu_H = \mu_{[1]} - (1-2/K)\Delta$ and $\mu_L = \mu_{[1]} - \Delta$.

\end{document}